\documentclass{article}

\usepackage{arxiv}

\usepackage[utf8]{inputenc} 
\usepackage[T1]{fontenc}    
\usepackage{hyperref}       
\usepackage{url}            
\usepackage{booktabs}       
\usepackage{amsfonts}       
\usepackage{nicefrac}       
\usepackage{microtype}      
\usepackage{lipsum}
\usepackage{graphicx}
\graphicspath{ {./images/} }

\usepackage{xyling} 
\usepackage[linguistics]{forest} 
\usepackage{bm}
\usepackage{amsmath}
\usepackage{biblatex}
\usepackage{tipa}
\title{Reliable detection and quantification of selective forces in language change}

\author{
Juan Guerrero Montero\\
    SUPA, School of Physics and Astronomy\\ University of Edinburgh\\
  \texttt{J.A.Guererro-
Montero@sms.ed.ac.uk} \\
   \And
 Andres Karjus\\
  ERA Chair for Cultural Data Analytics, Tallinn University\\
  School of Humanities, Tallinn University\\\\
  \texttt{andres.karjus@tlu.ee} \\
  \And
 Kenny Smith\\
  School of Philosophy, \\Psychology and Language Sciences \\
  University of Edinburgh \\
  \texttt{kenny.smith@ed.ac.uk} \\
  \AND 
Richard A. Blythe\\
  SUPA, School of Physics and Astronomy\\ University of Edinburgh\\
  \texttt{r.a.blythe@ed.ac.uk}
}

\begin{document}
\maketitle
\begin{abstract}
Language change is a cultural evolutionary process in which variants of linguistic variables change in frequency through processes analogous to mutation, selection and genetic drift. In this work, we apply a recently-introduced method to corpus data to quantify the strength of selection in specific instances of historical language change. We first demonstrate, in the context of English irregular verbs, that this method is more reliable and interpretable than similar methods that have previously been applied. We further extend this study to demonstrate that a bias towards phonological simplicity overrides that favouring grammatical simplicity when these are in conflict. Finally, with reference to Spanish spelling reforms, we show that the method can also detect points in time at which selection strengths change, a feature that is generically expected for socially-motivated language change. Together, these results indicate how hypotheses for mechanisms of language change can be tested quantitatively using historical corpus data.
\end{abstract}

\section{Introduction}

Human languages undergo constant change as a result of innovations in form and meaning, and  competition between new and old forms of expression. For example, a phoneme may start being pronounced in a different way, or a new word order may be introduced. A wide range of factors may be responsible for innovation \cite{ref:Variation}. These include expressivity, for example, a desire to be noticed, recognisable, amusing, or charming \cite{ref:KellerInvisibleHand}, and economy, minimizing the effort needed to communicate without compromising the listener's understanding, which may lead to the development of novel, simpler forms \cite{ref:ZipfEconomyPrinciple}. Meanwhile, competition mediated by interactions with other components of the language may favour a more consistent mapping between form and function. Alongside these, social factors like prestige or taboo \cite{ref:LabovLinguisticChangeSocial}, may make certain variant forms more or less attractive to certain language users.

In this work, our aim is to quantify the competition between linguistic variants that are available to a speech community, and therewith gain insights into its origins. We achieve this by viewing language change as a cultural evolutionary process \cite{ref:CroftLanguageChange,ref:Mufwene,ref:AtkinsonGray,ref:Pagel2009}. When modelling cultural evolution \cite{ref:Cavalli,ref:BoydRicherson}, it has long been recognised that changes in variant frequencies may arise both from systematic biases (which we refer to generically as \emph{selection}) and random  \emph{drift}. While drift may refer to directional change in linguistics following \cite{Sapir}, we use it here in the cultural evolutionary sense, denoting unbiased stochastic change. Typically one is most interested in identifying the selective forces that cause one variant to be favoured over another, including linguistic \cite{ref:LabovLinguisticChangeInternal} or social \cite{ref:LabovLinguisticChangeSocial} factors. Eliminating the possibility that changes may be entirely due to drift is a necessary first step in this endeavour. Initial attempts to achieve this in the context of cultural evolution involved establishing statistical properties of drift and comparing with the corresponding features of empirical data. For example, the distributions of baby names \cite{ref:DriftBabyNames} and Hittite ceramic bowl types \cite{ref:Steele}, as measured at a single point in time, were found to be consistent with the predictions of drift. Under closer examination, however, deviations from drift were found in both cases, for example, by appealing to the rate at which the most abundant types are replaced \cite{ref:AcerbiBentley}.

Cultural and linguistic datasets provide a potentially rich source of data to constrain parameters in a model of the evolutionary process. In particular, by combining observations of token frequencies at multiple time points, one should achieve greater inferential power than can be achieved by considering only a single point in time. Although such analyses are challenging to construct, a number of forward steps have been made in recent years. For example, the evolution of pottery styles was investigated by appealing to predictions for the number of types remaining after a given time under drift \cite{ref:KandlerShennan} and by using simulated trajectories of variant frequencies in an Approximate Bayesian Computation scheme \cite{ref:KandlerPowell}.

Here, we analyse changes in linguistic corpus data with a method based on the Wright-Fisher model of evolution \cite{ref:Fisher,ref:Wright}. Although introduced as a model for changes in gene frequencies through biological reproduction, the Wright-Fisher model is also relevant to cultural evolution \cite{ref:Cavalli}. In the specific context of language change, the Wright-Fisher model has been shown to be equivalent to a variety of different conceptual approaches. For example, a mathematical formulation of Croft's descriptive theory of utterance selection \cite{ref:CroftLanguageChange}, itself grounded in \cite{ref:Hull}'s generalised analysis of selection, was shown to have the same structure as the Wright-Fisher model \cite{ref:UtteranceSelection}. Moreover, \cite{ref:WordsAsAlleles} showed that a version of the Wright-Fisher model that includes innovation and drift is equivalent to a model of iterated learning where language learners apply Bayesian inference to estimate a variant's frequency in their linguistic input. Other theories of language change, for example those that invoke a competition between multiple candidate grammars \cite{ref:YangGrammarComp}, can also be viewed as instances of Hull's generalised analysis of selection, and it has been argued that these may also be represented as a Wright-Fisher model \cite{ref:RichardPLOS}.

The essence of the analysis presented below is to determine the values of parameters in the Wright-Fisher model that maximise the probability that the model generates the series of variant frequencies obtained from a historical corpus. As we set out in Section~\ref{sec:methods} below, one of these parameters quantifies the strength of selection, and the other the scale of fluctuations arising from random contributions to language change. A difficulty with the Wright-Fisher model is that the mathematical formul\ae\ that describe the evolution are difficult to work with. In genetics, a great  deal of effort has been invested in devising reliable approximations that facilitate application to empirical time series \cite{ref:Tataru2}, an effort that we utilise here in the cultural evolutionary context. Specifically we build on a Beta-with-spikes approximation \cite{ref:BWS1} in a way that facilitates an efficient and reliable estimation of model parameters, as judged by benchmarking with both real and synthetic data \cite{ref:Us}.

In Section~\ref{sec:results} we apply this method to historical corpus data in three separate investigations. First, we revisit the set of English verbs with irregular past tense forms that were previously examined by \cite{ref:FITApplication}, \cite{ref:ProblemsofFIT} and \cite{ref:NeuralNetworks}, showing that our method is more reliable than that based on a normal approximation of the Wright-Fisher model \cite{ref:FITApplication} while offering greater interpretability than a neural-network based time-series classifier \cite{ref:NeuralNetworks}. In common with \cite{ref:FITApplication}, we find that some verbs appear to be irregularising over time.

By itself, the inferred strength of selection is not necessarily informative as to its underlying cause. Our second investigation demonstrates one approach by which such information can be gleaned. Specifically, we divide English verbs into two sets: those whose regular past tense form contains a repeated consonant, and those that do not. The former set is then subject to a conflict between the greater grammatical simplicity that would be gained by following the regular pattern and the greater phonological simplicity afforded by omitting the repeated consonant \cite{LebenOCP,StermbergerHaplology}. By comparing the selection strengths between the two sets, we can show that the latter constraint tends to override the former in the context of English verbs.

Finally, we turn to a set of Spanish words that were affected by orthographical reforms in the 18\textsuperscript{th} and 19\textsuperscript{th} centuries.  Here, we demonstrate that an unsupervised maximum-likelihood analysis can pinpoint with good accuracy the time at which the reforms were introduced and furthermore quantify the impact of the reform on the linguistic behaviour of the speech community. These last results illustrate that, even with time-series comprising a few measurement points, we can uncover social changes that might not otherwise be apparent. We discuss such opportunities, along with limitations of our method, further in Section~\ref{sec:disco}.

\section{Methodology}\label{sec:methods}

\subsection{Maximum likelihood estimation methods}

Maximum likelihood estimation is a conceptually simple yet powerful technique for estimating parameter values in a model and selecting between multiple candidate models. The basic setup is that we have both some empirical data, denoted $X$, and a probabilistic model that tells us how likely the observation $X$ is given some choice of parameters $\bm{\Theta}$. We then estimate the values of the parameters by determining the combination that maximises the likelihood of the data. This procedure lies at the heart of many statistical methods, including linear regression. In such a model, parameters are chosen to maximise the likelihood of the data given a statistical model of the residuals \cite{ref:LikelihoodMethods1,ref:LikelihoodMethods2}, for example, that the residuals are drawn from a normal distribution. It can also be viewed as a special case of Bayesian inference with a uniform prior.

In this work we are concerned with frequency time series, that is, a sequence of measurements $X=\{(x_t,t)\}=\{(x_1,t_1),(x_2,t_2),...,(x_m,t_m)\}$ where $x_i$ is the fraction of instances of use of a linguistic variable (e.g., all past-tense forms of a specific verb) in which a particular variant (e.g., the regular form) was used during a short time window centred on time $t_i$. Thus, the dataset $X=\{(0.2,\,1),\,(0.5,\,2),\,(0.75,\,5)\}$ would imply a proportion of usage of the regular form of $20\%$ at time $1$, $50\%$ at time $2$ and $75\%$ at time $5$ (and no frequency data at any other time).

The underlying evolutionary model of language change determines a set of transition probabilities, $\text{Prob}(x_{i+1},t_{i+1}|x_{i},t_{i},\bm{\Theta})$, that tell us how likely it is that, given a proportion $x_i$ at time $t_i$ and parameters $\bm{\Theta}$, the proportion will be $x_{i+1}$ at time $t_{i+1}$. In the previous example, the dataset $X$ would determine the transition probabilities $\text{Prob}(0.5,2|0.2,1,\bm{\Theta})$ and $\text{Prob}(0.75,5|0.5,2,\bm{\Theta})$, whose exact numerical values would depend on the choice of model parameters $\bm{\Theta}$. We assume that contributions to changes in variant frequencies at different points in time are uncorrelated, which means that we can write the likelihood of the entire frequency time series as the product of the transition probabilities for each interval:
\begin{equation}\label{eq:Likelihood}
    L(X|\bm{\Theta})=\prod_{i=1}^{m-1}\text{Prob}\left(x_{i+1},t_{i+1}|x_i,t_i,\bm{\Theta}\right) \,.
\end{equation}
It is this likelihood function that we will maximise to determine the set of parameters $\bm{\Theta}$ that best describes the cultural evolutionary dynamics, and that we will use to compare different models.

There are two main ways to choose the form of the transition probabilities $\text{Prob}(\cdot|\cdot,\bm{\Theta})$, a choice that is crucial to parameter estimation and subsequent interpretation. One could simply assume that the frequencies follow a prescribed trajectory between the two points, subject to some fluctuations around them. For example, in linear regression, frequencies would be assumed to vary linearly with time with normally-distributed residuals. Logistic regression is similar, but instead assumes that frequencies vary following a nonlinear logistic function that is commonly used as a model for S-shaped language change \cite{ref:sshapekroch_1989,ref:WordsAsAlleles,ref:KauhanenCRE}. A weakness of this approach is that without an underlying model of language production and transmission that may be operationalised as frequency time series, it is difficult to relate the parameters obtained to the behaviour of individuals or speech communities.

The alternative is to derive the transition probabilities from an explicit agent-based model of language change, many of which can be understood as a variant of the Wright-Fisher model of evolution \cite{ref:RichardPLOS}. As noted in the introduction, the transfer of a model from genetics is justified on theoretical grounds \cite{ref:Hull,ref:CroftLanguageChange} and one can interpret the parameters by appealing to models of language use \cite{ref:UtteranceSelection} or iterated Bayesian learning \cite{ref:WordsAsAlleles}. A drawback of the Wright-Fisher model is that exact expressions for the transition probabilities \cite{ref:WFBook} are complex and difficult to work with computationally, as their associated transition matrices may become numerically intractable \cite{ref:BwSComparisonTechnicalParis}. This has motivated many different approximation schemes \cite{ref:Tataru2}. In this work, we apply a self-contained Beta-with-Spikes approximation scheme that was developed and tested by \cite{ref:Us} and found to provide reliable estimates for parameter values without incurring undue computational cost. In the following we overview the conceptual components of this approach that are most relevant to linguistic applications, directing the reader to \cite{ref:Us} for technical details.

\subsection{The Wright-Fisher model}\label{sec:wf}

\begin{figure}[tbp]
\begin{center}
\scalebox{0.25}{\includegraphics{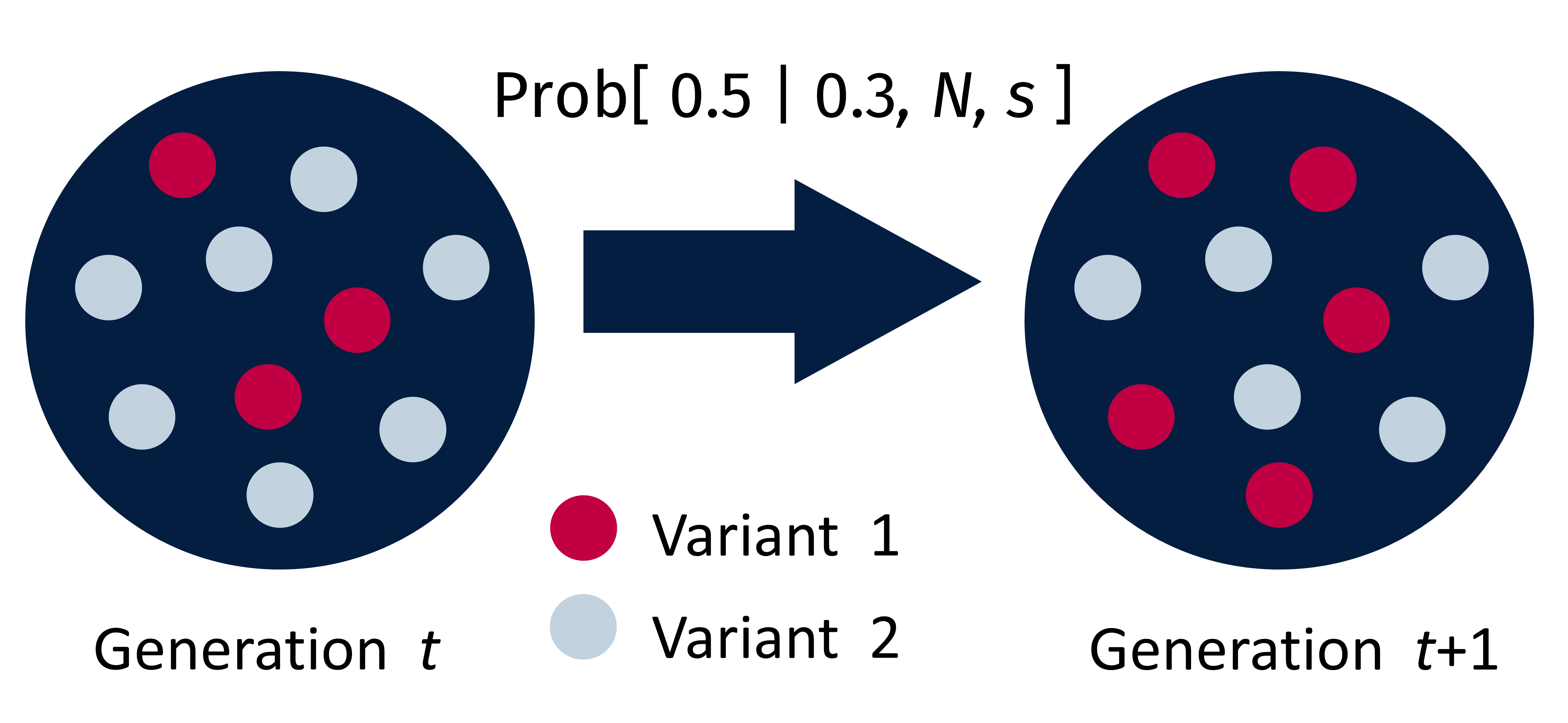}}
\end{center}
\caption{Schematic representation of the transition from generation $t$ to generation $t+1$ in a Wright-Fisher process with $N=10$.}
\label{fig:WFScheme}
\end{figure}

The Wright-Fisher model describes a population of $N$ replicating individuals of different types, each of which directly corresponds to a different variant of a linguistic variable. While any number of mutually exclusive types can be included in the model, we will focus on the case with two distinct types. Its extension to three or more distinct types is straightforward \cite{ref:Tataru2}. The quantity $x_t$ is the proportion of individuals of a specific variant in generation $t$, as described above. The process of replication has the effect that, in generation $t+1$, each individual is of the variant of interest with probability $g(x_t,s)$, which depends both on the composition of the population and a measure of selection strength that we denote $s$. It is assumed that each of the $N$ individuals in the new generation has its type assigned independently. This replication process is illustrated in Figure~\ref{fig:WFScheme}.

The two evolutionary parameters, $N$ and $s$, can be afforded a linguistic interpretation as follows. The effective population size, $N$, quantifies the scale of fluctuations in the variant frequencies around a smooth trajectory of change. The smaller the effective population size, $N$, the larger the fluctuations. When selection is weak ($s$ is close to zero), the time taken for a variant to go extinct is proportional to $N$ \cite{ref:WFBook}. Its interpretation as a population size, effortless in population genetics, does not work as well when studying language change. Through agent-based models of language learning and use \cite{ref:UtteranceSelection,ref:RichardPLOS}, we understand that $N$ generically correlates with the size of the speech community. However, heterogeneous social network structures can result in $N$ correlating only weakly with the size of the human population \cite{ref:Bromham,ref:Wichmann,ref:RichardPLOS}. The population size $N$ is also related to the total usage of all variants of a linguistic variable in a given generation, and how long it is retained in memory \cite{ref:UtteranceSelection,ref:WordsAsAlleles}. Intuitively, speakers will be more consistent in their usage of specific variants the more they encounter them, and the longer they recall these encounters. This increased individual consistency will be reflected as smaller fluctuations in time series data. Although our analysis will not allow these different contributions to $N$ to be distinguished, we will be able to determine which linguemes are subject to greater or lesser uncertainty in transmission between speakers.

The selection strength parameter, $s$, represents a tendency for the variant of interest to increase in relative usage ($s>0$) or decrease ($s<0$). Here, $s$ subsumes all factors that could lead to a variant systematically increasing or decreasing in frequency over time, whether they originate in cultural, cognitive or language internal factors \cite{ref:LabovLinguisticChangeInternal,ref:LabovLinguisticChangeSocial,ref:LabovLinguisticChangeCognitiveCultural,ref:CroftLanguageChange}. Similarly to the various factors that may influence the effective population size, we will not be able to distinguish them from the value of $s$ alone. However, as we show below, we can gain useful information by looking for common features of variants which are found to have similar selection strengths.

\begin{figure}[tbp]
\begin{center}
\includegraphics[width=0.7\linewidth]{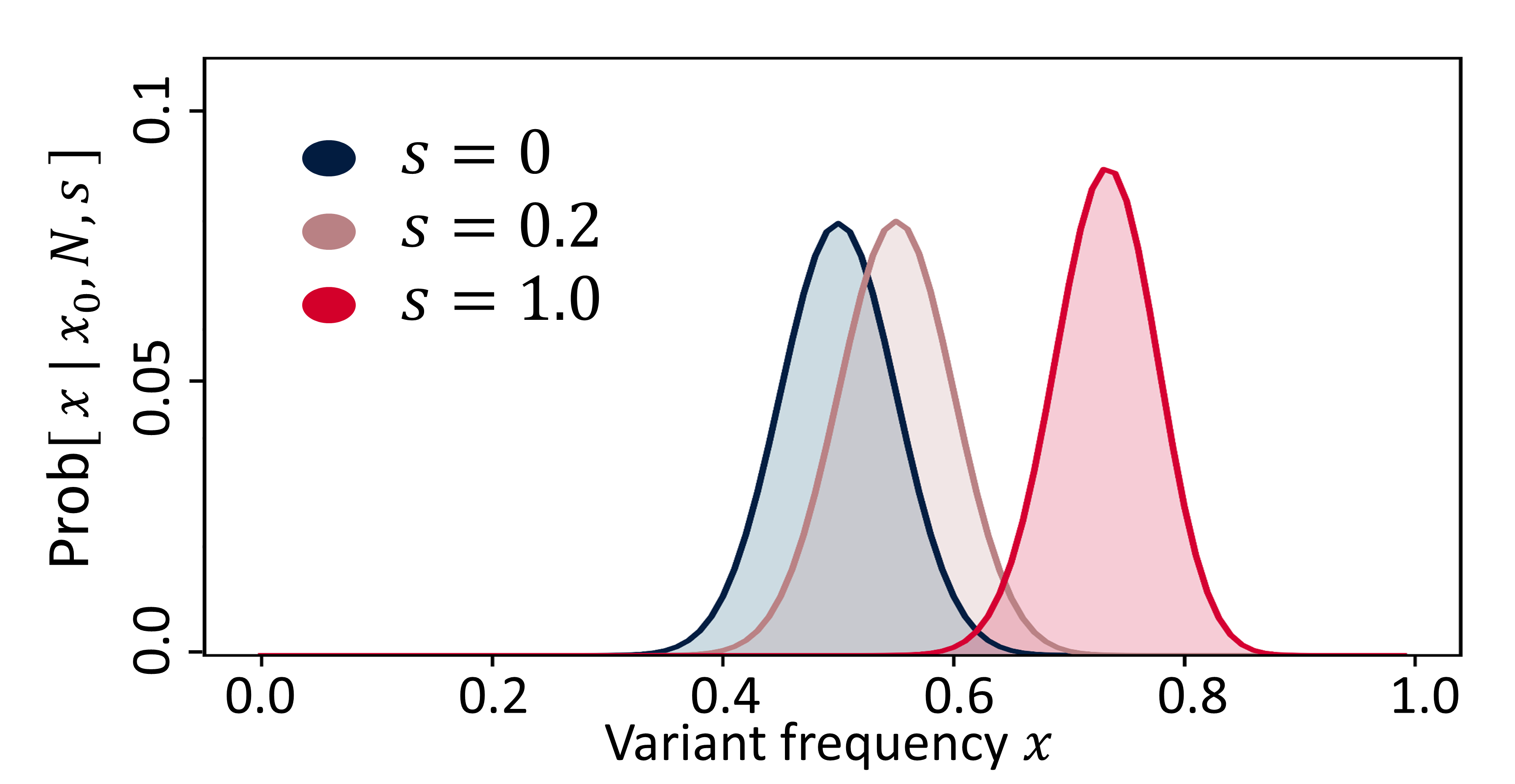}
\end{center}
\caption{Probability distribution of a variant frequency $x$ after one generation of evolution in the Wright-Fisher model, starting from a frequency $x_0=\frac{1}{2}$. As the selection strength $s$ increases, the distribution becomes more sharply peaked on larger values of $x$.}
\label{fig:DifferentS}
\end{figure}

The parameter $s$ specifies the probability $g(x,s)$ that an individual in generation $t+1$ is an offspring of an individual with frequency $x$ and selection strength $s$ in the previous generation. Here, we take
\begin{equation}\label{eq:s}
    g(x,s)=\frac{1}{1+\frac{1-x}{x}e^{-s}} ,
\end{equation}
which has been commonly used in the theoretical characterisation of language change \cite{ref:KauhanenCRE,ref:YangAcquisition}. In Figure~\ref{fig:DifferentS} we plot the transition probability $\text{Prob}(x|x_0,s)$ that results from this definition for the case where $x_0=\frac{1}{2}$ and for different values of $s$. We see that larger values of $s$ shift the peak of this distribution towards higher values of this frequency $x$, consistent with the notion of a bias towards the corresponding variant.

In the literature, one can find relationships between the selection strength $s$ and the probability $g(x,s)$ different to that specified above \cite{ref:WFBook,ref:Tataru2}. Our chosen formula has the useful property that $g(x,s)+g(1-x,-s)=1$, which means that if one of two variants in a population has a selection strength $s$, the other one implicitly has a selection strength $-s$. This choice thus lends a symmetry between positive and negative selection strengths of the same magnitude, which aids the interpretation of the results. The strength $s=0$ represents pure drift, where any changes in usage over time are due to the stochasticity of replication alone, and not the presence of selective forces. Under pure drift, $g(x,s)$ reduces to $g(x,0)=x$.

\subsection{Beta-with-Spikes approximation}\label{sec:bws}

For a single generation of evolution in the Wright-Fisher model, the transition probability is the binomial distribution
\begin{equation}\label{eq:WFTransitionProb}
    \text{Prob}(x_{t+1}|x_{t},N,s)=
    \begin{pmatrix}
    N\\
    Nx_{t+1}
    \end{pmatrix}
    g\left(x_{t},s\right)^{Nx_{t+1}}\left(1-g\left(x_{t},s\right)\right)^{N(1-x_{t+1})}
\end{equation}
because there are $N$ individuals and a success probability of $g(x_t,s)$.
If the effective population size $N$ is known, and the time between two frequency measurements corresponds to a single generation, one can use this expression for the transition probability in Equation~(\ref{eq:Likelihood}) to construct the overall likelihood of a series of measurements. In the present application to linguistic corpus data, neither of these requirements hold. $N$ is a parameter that we need to estimate, and measurement times are not in general separated by a fixed interval that constitutes a single Wright-Fisher generation. The Beta-with-Spikes approximation, introduced by \cite{ref:BWS1} and extended by \cite{ref:Us}, is designed to deal with these complexities.

For two observations made at times $x_t$ and $x_{t+k}$ (i.e., separated by $k$ generations) the BwS approximation is
\begin{equation}\label{eq:BWS}
\begin{split}
    \text{Prob}(x_{t+k}|x_t,N,s) &=P_{0,k}\delta(x_{t+k})+P_{1,k}\delta(1-x_{t+k})\\
    &+\left(1-P_{1,k}-P_{0,k}\right)\frac{x_{t+k}^{\alpha_k-1}(1-x_{t+k})^{\beta_k-1}}{\text{B}(\alpha_k,\beta_k)} \,.
\end{split}
\end{equation}
Here, $P_{0,k}$, $P_{1,k}$, $\alpha_k$ and $\beta_k$ are parameters that determine the shape of the distribution. These parameters have the following interpretation. $P_{0,k}$ is the probability that the variant has gone extinct by the $k^{\rm th}$ generation, and $P_{1,k}$ is the probability that it has driven the other variant to extinction in that time. $\alpha_k$ and $\beta_k$ control the shape of the distribution of variant frequencies, conditioned on neither of them having gone extinct. These parameters can be determined from the mean and variance of this conditional distribution (see \cite{ref:Us}). Note that all four parameters depend on $N$ and $s$, as well as the sequence of observed frequencies $x_{t_i}$. Therefore, they need to be recalculated for each time series and combination of model parameters.

\begin{figure}[tbp]
\begin{center}
\scalebox{0.3}{\includegraphics{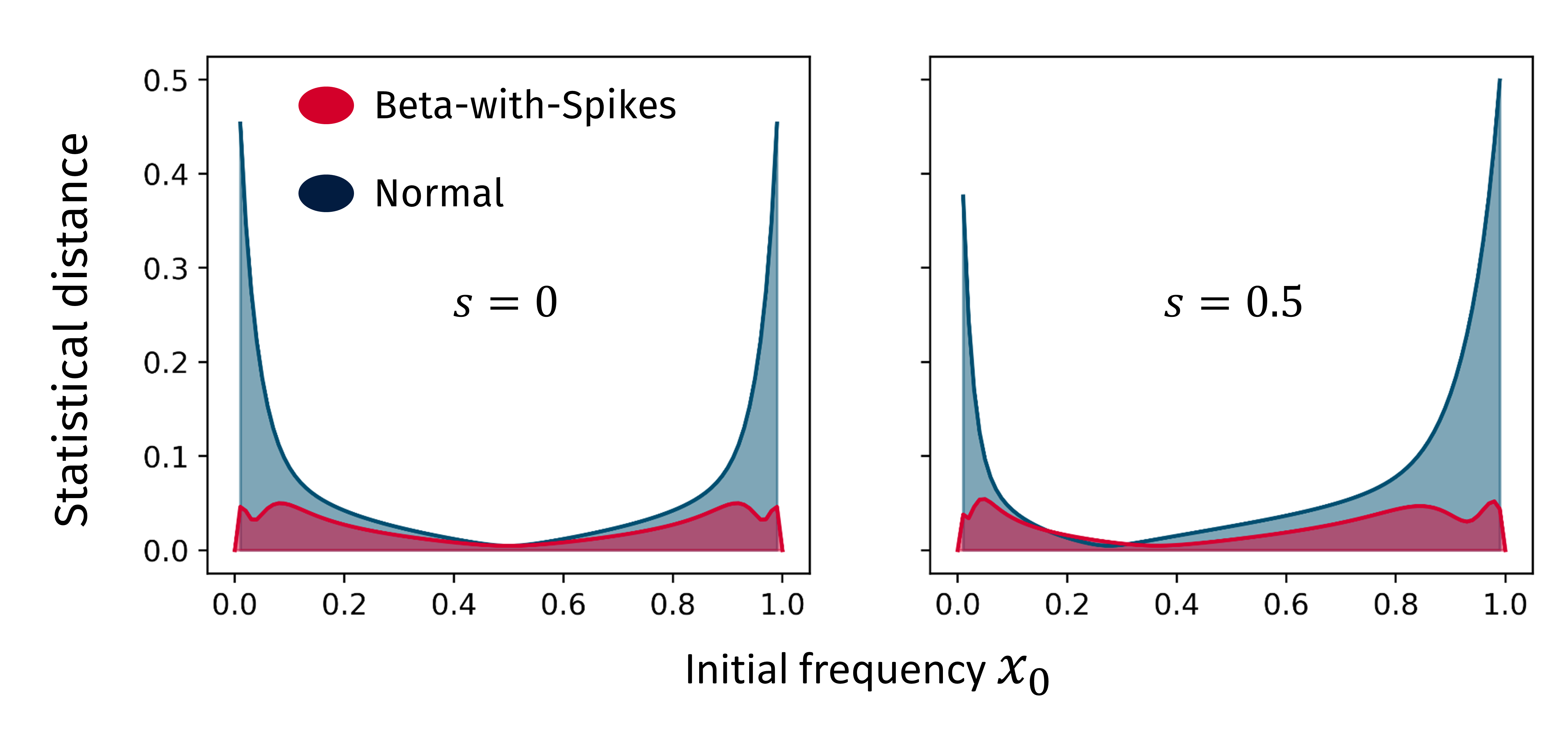}}
\end{center}
\caption{Comparison of the statistical distances of the BwS and normal approximations to the exact WF distribution as a function of the initial frequency $x_0$. Left: statistical distance for pure drift ($s=0$). Right: statistical distance for strong selection ($s=0.5$). The Beta-with-Spikes approximation has lower statistical distance to the exact distribution (meaning it approximates it more accurately) for every value of $s$ and $x_0$, but especially for extreme values of $x_0$ close to $0.0$ or $1.0$ and for strong selection.}
\label{fig:BwSComparison}
\end{figure}

A crucial advantage of the BwS approximation is that it accounts for the fact that changes in variant frequencies cannot be arbitrarily large. If a variant has a low frequency ($x$ close to zero), then a downward fluctuation should cause it to become extinct, rather than attain a negative frequency. It is the spikes (represented by the delta functions) in the Beta-with-Spikes expression (\ref{eq:BWS}) that incorporate this constraint. By contrast, a normal approximation to the same transition rates (as used by \cite{ref:FIT,ref:FITApplication}) allows, in principle, arbitrarily large or negative $x$, instead of being constrained to the range $0\le x \le 1$.  This difference is illustrated in Figure~\ref{fig:BwSComparison} which shows the statistical distances between  the BwS and normal approximation and the exact WF transition probability, for different values of the initial frequency $x_0$ and two values of the selection strength $s$. We see that for both pure drift ($s=0$) and strong selection ($s=0.5$), the BwS approximation stays consistently closer to the exact distribution for all values of $x_0$, which is reflected in lower values of the statistical distance. In particular, the BwS approximation is significantly better than the normal approximation for initial frequencies $x_0$ close to the edges of the interval, and for strong selection.

The main task in applying the BwS approximation is to estimate the parameters $P_{0,k}$, $P_{1,k}$, $\alpha_k$ and $\beta_k$ for successive generations $k=1,2,3,\ldots$. The strategy of \cite{ref:BWS1} is to match up the moments of the BwS distribution to those of the Wright-Fisher model after $k$ generations have elapsed. This method works well when the selection strength $s$ small, but less so when it is large. \cite{ref:Us} have improved on the method, particularly in the large $s$ regime, by iterating (\ref{eq:WFTransitionProb}) one generation at a time, and reading off the extinction probabilities, mean and variance at each. The code that implements this procedure, and generates parameter estimates is available \href{https://osf.io/qxgnj/?view_only=de983f003307492bb6dd777ee3e36a39}{here}.

In the context of cultural evolution, it is not obvious what period of time counts as a generation in the Wright-Fisher model. In principle, this is a free parameter which would also need to be optimised by maximum likelihood estimation (and furthermore demand an interpretation). Fortunately, this is unnecessary. \cite{ref:Us} further show that the optimum values of $1/N$ and $s$ are both proportional to the chosen generation time. In other words, the generation time serves only to set the units in which the parameters $N$ and $s$ are measured. It is however important to use the same generation time across multiple time series when one wishes to compare the values of $N$ and $s$ that are obtained: otherwise, they would be in different units and not comparable. In this work we generally take the shortest time between successive observation points as the generation time. If one makes it shorter than this, the computational effort increases without any improvement in the quality of the estimates obtained. If one makes it longer, one must then aggregate multiple data points which then entails a loss of temporal resolution. However, as we discuss below, it is sometimes beneficial to combine data points to reduce sampling error that is not accounted for in the present maximum likelihood analysis.

\subsection{Distinguishing selection from drift}\label{sec:Method1}

As established in the introduction, the social, linguistic and cognitive forces driving language change are very diverse. Still, their measurable effects can be broadly characterised as belonging to one of two types. Systematic biases drive the evolutionary process in a specific direction, and can be modelled as selective forces. Frequency effects and stochasticity in transmission produce random, unbiased drift whose effects are always present, albeit not always sufficient to explain the behaviour of the data. Quantitative, empirical analyses benefit from the simple yet powerful and flexible characterisation of language change afforded by this binary description.

By using the transition probabilities (\ref{eq:BWS}) in the likelihood function (\ref{eq:Likelihood}), we can find the maximum likelihood values of the effective population size $N^*$ and selection strength $s^*$ via
\begin{equation}
    (N^*,s^*)=\text{arg max}\,L(X|N,s) .
\end{equation}
In practice, we find that the likelihood function $L(X|N,s)$ has a single maximum, which can be located by successively optimising on $N$ at fixed $s$ and vice versa.

It is important to establish whether the selection strength $s^*$ is significantly different to zero: otherwise, the null model of stochastic drift ($s=0$) would be sufficient to explain the behaviour of the data without the need for selection \cite{ref:DriftasNullHypothesis,ref:FITApplication}. In order to do this, we compare the maximal likelihood under selection, $L(X|N^*,s^*)$, with the maximal likelihood under pure drift. That is, we first restrict to $s=0$ and determine the optimal effective population size $N^*_0$:
\begin{equation}
    N^*_0=\text{arg max}\,L(X|N,0) \,.
\end{equation}
Then we compare the models with and without selection by computing the likelihood-ratio \begin{equation}
    \lambda=2\ln\left(\frac{L(X|N^*,s^*)}{L(X|N_0^*,0)}\right)\,.
\end{equation}
This quantity can be compared to a reference distribution to find a $p$-value, an estimation of the probability that the observed time series could have arisen from drift alone \footnote{The commonly used $\chi^2$ distribution does not work well for $p$-value estimation when working with data from historical corpora, as it only converges to the true distribution of likelihood-ratios for time series of infinite length \cite{ref:FIT}.} \cite{ref:LikelihoodMethods1,ref:LikelihoodMethods2}. To achieve this, we follow the procedure outlined by \cite{ref:FIT} and generate 1,000 artificial time series spanning the same time period as the empirical data $X$ with parameter values $s=0$ and $N=N_0^*$. For each of these we compute the maximum likelihood values $N^*$, $s^*$ and $N_0^*$, using the same sequence of steps as for the original empirical time series. We then compute the likelihood ratio $\lambda$ and determine what fraction of the artificial time series has a larger $\lambda$ than the one that was observed. This provides an empirical $p$-value for the null hypothesis of drift.

\section{Results}\label{sec:results}

We now apply the methods set out above in three separate tasks, each with a distinct purpose. First, we revisit the set of verb time series from the Corpus of Historical American English (COHA, \cite{ref:COHA}) to benchmark our approach against those of \cite{ref:FITApplication} and \cite{ref:NeuralNetworks}. These results demonstrate that the BwS method is both more robust than a similar likelihood-based approach \cite{ref:FITApplication} and more informative than a neural network trained to perform a binary classification \cite{ref:NeuralNetworks}. We also introduce a method for assessing the variability of parameter values under different binning strategies, thereby facilitating a judgement as to which results are more robust.

We then perform similar analyses to understand the direction of selection in the context of English irregular verbs, this time using the English 2019 1-grams and 2-grams datasets from the Google Books corpus \cite{ref:GoogleBooks}. This larger corpus contains more instances of verbs that appear to be irregularising over time. We find that a phonological constraint that disfavours repeated consonants can override a general preference for regularity. Finally, we use data from the 2019 Spanish 1-gram Google Books corpus to show that the dates at which Spanish spelling reforms were introduced can be detected using the unsupervised maximum-likelihood analysis.

The validity of using frequency data from the Google Books corpus to draw conclusions on cultural evolution and language change has been questioned by \cite{ref:PechenickProblemswithGoogleBooks} due to the over-representation of scientific literature in the English sub-corpus throughout the 20th and 21st centuries. While they propose restricting studies of cultural and language change to the fiction sub-corpus, we believe that using frequency data from the general English sub-corpus is justified for the purposes of our study. First, our work rests on the comparison between two data sets of English verbs differing only in their phonology. It is reasonable to assume that, if any bias exists in scientific texts regarding the use of irregular or regular forms of verbs, this bias will not be phonologically conditioned, thus maintaining the validity of the comparison between both data sets. Secondly, we have chosen verbs that are reasonably present in both the general English corpus and the English Fiction corpus, so a potential biases towards uncommon verbs in scientific literature should not be an issue. Thirdly, the general English sub-corpus will contain more words than the restricted fiction sub-corpus, thus reducing the effect of sampling noise on our results.

\subsection{Drift vs selection in past-tense English verbs}\label{sec:COHA}

A simple example of competition between two variants is provided by English verbs with an irregular past tense form which in many cases coexists with a regular form. This competition has been studied from a variety of quantitative perspectives \cite{ref:Lieberman,ref:Cuskley,ref:FITApplication,ref:ProblemsofFIT,ref:NeuralNetworks,YangTolerancePEnglish}. Of greatest relevance to the present work are those studies that aimed to distinguish drift from selection as the mechanism behind changes in the relative frequencies of the regular and irregular forms over time.

\cite{ref:FITApplication} applied the Frequency Increment Test (FIT, \cite{ref:FIT}) to a set of verbs from the Corpus of Historical American English (COHA, \cite{ref:COHA}). This is a maximum-likelihood method that rests on a normal approximation to the Wright-Fisher transition probabilities. Like the Beta-with-Spikes maximum-likelihood method in Section \ref{sec:Method1}, this method yields estimates of the effective population size and selection strength, along with a $p$-value for the null hypothesis of pure drift. However, there are situations where results are flagged as unreliable due to the frequency increments failing a normality test \cite{ref:FITApplication}. \cite{ref:ProblemsofFIT} further noted that the results can also be sensitive to the size of the window over which frequencies are estimated.

\cite{ref:NeuralNetworks} avoid these issues by taking the rather different approach of training a neural network on simulated time series generated by the Wright-Fisher transition probabilities (\ref{eq:WFTransitionProb}) for different values of $s$ (but fixed $N=1,000$). Each time series in the training set is labelled according to whether it was generated purely by drift ($s=0$) or if selection was operating ($s\ne0$). Once trained, the network yields a binary classification of empirical time series, according to whether they are more similar to the examples of drift or selection in the training data. We refer to this as the Time Series Classification (TSC) approach. The advantage of TSC is that no approximation to the Wright-Fisher transition probabilities is made. Moreover, one can manipulate the training data so that it displays artifacts of binning or finite sample sizes that are features of real time series, which in turn should improve the reliability of the classification. This approach does however come with some drawbacks. Whilst the output from the classification algorithm is a value between 0 and 1, it does not have an obvious interpretation as a probability. \cite{ref:NeuralNetworks} used a threshold of $0.5$ to label timeseries as arising from drift or selection. The method further does not provide an estimate of the strength of $s$, and since $N$ was fixed in the training set, this amounts to an assumption that this single value of $N$ was appropriate for all empirical time series.  This could be an issue since \cite{ref:FITApplication} report a wide range of values of $N$ for this data set (from around $80$ to around $22,500$).

\begin{figure}[tbp]
    \begin{center}
    \includegraphics[width=\linewidth]{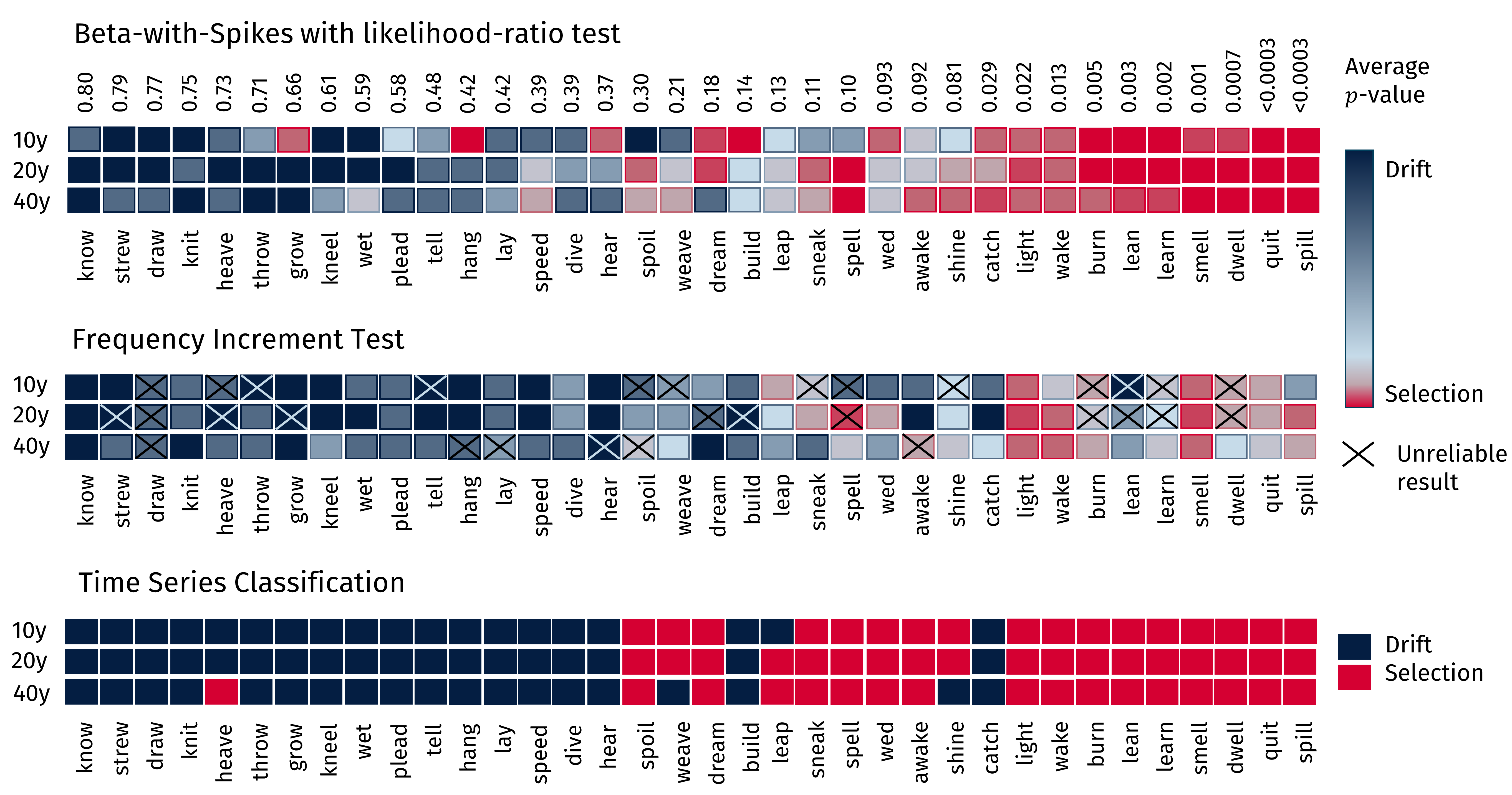}
    \end{center}
    \caption{
    Results for the detection of selective forces in 36 COHA verbs, with three different methods and for three different temporal binnings of 10, 20 and 40 years. Results for both the FIT and BwS likelihood-ratio algorithms produce a $p$-value for the pure drift hypothesis. Blue shades represent higher $p$-values (i.e., higher likelihood of the data under drift), while red shades represent $p$-values under the traditional $0.05$ threshold of significance for selection. Time series where the normal approximation that FIT relies on is inaccurate are crossed out. Results for the TSC method from \protect\cite{ref:NeuralNetworks} are classified in a binary way as either drift or selection. The average $p$-value across the three bins widths obtained through the BwS algorithm is shown along the horizontal axis. We note that the BwS method gives results consistent with TSC when FIT is unreliable.}
\label{fig:COHAResults}
\end{figure}

In Section A of the appendix we report the maximum likelihood estimates of $N$ and $s$, along with the $p$-value for the drift hypothesis, obtained using the BwS method for the same set of verbs that were considered by \cite{ref:FITApplication} and \cite{ref:ProblemsofFIT} using FIT and by \cite{ref:NeuralNetworks} using TSC. We perform the analysis by extracting annual frequency data of the variants of interest from COHA and aggregating it into 10-, 20- and 40-year bins. The reason for this is a trade-off between the more precise frequency estimates that derive from larger bins and the greater temporal resolution obtained from a larger number of bins over the relevant historical period. By employing different binning strategies, we can gain insights into the consequences of this trade-off. Variable-width binning strategies have also been successfully applied in previous studies \cite{ref:FITApplication}. In these, the number of tokens per bin is kept roughly constant at an arbitrarily chosen value, at the expense of varying their temporal width. For the purpose of comparing the different methods, we have chosen to look only at fixed-binning strategies, although the BwS method could be combined with variable-width binning.

We focus first on the role played by selective forces, which we quantify by appealing to the $p$ values associated with the null hypothesis of pure drift as described in Section~\ref{sec:Method1}. In Figure~\ref{fig:COHAResults} we compare the results obtained from the three different methods by ordering the verbs from left to right by decreasing BwS $p$-value, averaged over the three temporal binnings. Each panel corresponds to a different analysis method, and indicates the $p$-value for the hypothesis of pure drift for each verb and binning protocol. We recall that higher $p$-values are more suggestive of the historical changes being due to drift: these are represented with colours ranging from light to dark blue, with darker colours representing higher $p$-values.
Meanwhile, low $p$-values point towards other forces (such as selection) being present and are represented with different shades of red. 
While we use the standard $p$-value threshold of $0.05$ in the transition between blue (drift) and red (selection) in this representation, we acknowledge that these mechanisms lie in a continuum by making the transition between these extremes smooth.

We see from Figure~\ref{fig:COHAResults} that the three distinct methods give broadly consistent results, with those verbs towards the left being more compatible with change through pure drift, and those to the right with change from selection. More precisely, the correlation coefficients between the $p$-values obtained with different methods are 0.63 (Pearson) between FIT and BwS, 0.68 (biserial) between TSC and FIT, and 0.62 (biserial) between BwS and TSC. Analyses producing high $p$-values for selection (i.e. implying that drift alone can explain the behaviour of the data) are indicated with blue colours, whereas those where selection is more significant are red. Results obtained through the FIT method are generally consistent with those obtained with the BwS method. However, 30 of the FIT results (27.8\% of the total) are flagged as `unreliable' due to a failed normality test. These reliability issues are designed out of the BwS method, as it does not require normally-distributed increments \cite{ref:Tataru2,ref:Us}. Confidence in the method's reliability is also gained by benchmarking with synthetic and genetic data \cite{ref:Us} and through the consistency with the independent TSC results. The higher precision of the BwS at high selection strengths leads to higher significance (lower $p$-values) in its detection of selective forces when compared to the normal approximation, leading to redder colours in Figure~\ref{fig:BwSComparison}.

The TSC appears to give a cleaner classification of verbs according to drift and selection, and greater consistency with different choices of bin size. This is likely due in part to the training data being subjected to the same binning protocol as the empirical time-series, but also because a strict threshold was applied to the neural network's output value to partition into the two classes. While the TSC neural networks produce a value between $0$ and $1$ as their output, making it more nuanced than this binary classification would suggest, this number is not a probability or a $p$-value like those produced by BwS or FIT. Thus, an arbitrary threshold is necessary in order to classify time series as driven by drift or selection. A higher or lower threshold would put the boundary between the two classes in a different place. This hinders the interpretability of the result and the estimation of significance levels.

Our results further demonstrate that variation in $p$-values under different binning strategies, previously observed within the FIT analysis \cite{ref:ProblemsofFIT}, remains evident under the less restrictive BwS analysis presented here. We consequently regard this variability as an inherent feature of the time series data: that is, some changes are harder to classify than others. That is, this uncertainty need not be a failure of the method, but a reflection of linguistic reality. For example, it could reflect different variants being used less predictably by speakers, or by the constraints on variation changing over short timescales \cite{ref:TagliamonteSociolinguistics}.

\begin{figure}[tbp]
\begin{center}
\includegraphics[width=0.8\linewidth]{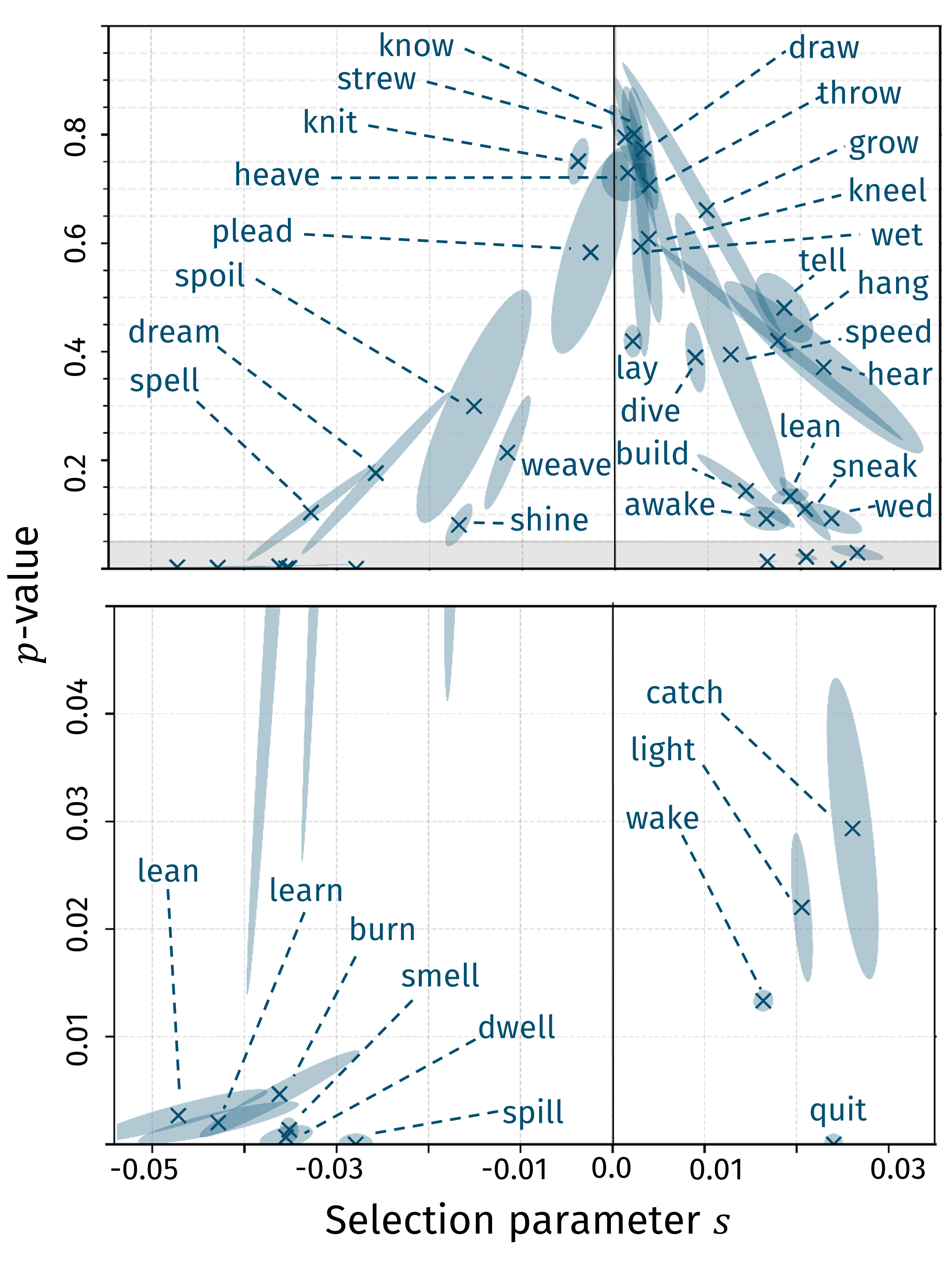}
\end{center}
\caption{Variability in the selection strengths $s$ and $p$-values for the null hypothesis pure drift for the COHA verbs. Each cross shows the mean value of the two parameters for each verb obtained when aggregating frequencies into temporal bins of different lengths. Each ellipse indicates the variability in the parameters at the level of one standard deviation. The vertical axis is an indicator of selection, defined as one minus the $p$-value associated with the drift hypothesis. The lower panel shows those verbs that fall within the range of $p$-values that is conventionally used to reject the null hypothesis for a single observation. In this panel we see a clear split into those that are regularising (negative $s$) and are irregularising (positive $s$).}
\label{fig:COHAEllipses}
\end{figure}

Such observations motivate a more detailed investigation of the classifiability of individual time series. A time series that shows limited variation in parameter values under different temporal binnings is more classifiable than one that shows more variation. With our interest in selection, the two most relevant parameters are $s$, the selection strength, and the $p$-value associated with the drift hypothesis. We can visualise the variation in these parameters by performing a Principal Components Analysis \cite{ref:PrincipalComponentAnalysis} on combinations obtained through different binning strategies (in this case, bins of 10, 20 and 40 years). 
The interior of the resulting ellipses indicates the range of variation of the two parameters over different binning strategies. This way, they provide a visualization of not just the average, but the uncertainty and covariance of $s$ and the $p$-value under different binning strategies. We show these ellipses for the COHA verbs in Figure~\ref{fig:COHAEllipses}. The upper panel contains the full range of $p$ and $s$ values obtained through the analysis, while the lower panel zooms in on the region where the drift $p$-value is smaller than $0.05$ (i.e., the conventional threshold for rejecting the null hypothesis). We see a correlation between the maximum likelihood value of $s$ and the $p$-value (both through the positions and rotation angles of each ellipse). 

The ellipses that lie entirely within the lower panel correspond to the verbs that are most likely to be driven by selection. We see a clear split between four verbs with positive selection (catch, light, wake and quit), which corresponds to them becoming more irregular over time, and six verbs (learn, lean, burn, smell, dwell and spill) with negative selection, and thus regularising over time. In this analysis, the frequency $x$ is the fraction of irregular forms used in the relevant context. Across the entire plane, there is evidence of both regularisation and irregularisation, although in most cases it is difficult to rule out drift as an explanation for the changes, as was observed by \cite{ref:FITApplication}. 

In interpreting these results, it is important to recognise that the presence of fluctuations around a smooth change trajectory will tend towards a higher drift $p$-value, since in the analysis drift is the sole source of fluctuations. It is possible that fluctuations in the corpus derive from other sources, such as sampling effects associated with a finite corpus. Some methods for estimating parameters in the Wright-Fisher model attempt to account for such fluctuations separately to drift \cite{ref:Tataru2}. These are, however, typically difficult to implement, and instead we sidestep the issue by ensuring sufficiently many tokens in each temporal bin that the frequency is well estimated. As such, we might expect to see stronger evidence for selection as the bin width is increased, which appears to be true for some (but not all) of the verbs with intermediate $p$-values. This suggests that some language changes may be dominated by the random effects of drift and therefore exhibit strong fluctuations even in very large corpora.

To summarise, we have shown in this section that the BwS method can be readily applied to historical corpus data for changes in the frequencies of linguistic variants. It provides estimates of parameters in the Wright-Fisher model that do not rest on an assumption that frequency increments are drawn from a normal distribution, and we find broad consistency in the strength of support for a drift hypothesis with complementary methods.

\subsection{Competing linguistic motivations in English verbs}\label{sec:GBEnglishResults}

In the previous section we observed a split between some verbs that were regularising and some that were irregularising. While the extension of regular inflection at the expense of irregulars seems to be the norm (e.g.~\cite{bybee95regular, simswilliams_16_treatment}), irregularisation is however an attested phenomenon. \cite{ref:Cuskley} found that the processes of regularisation and irregularisation tend to take place with similar frequency,
something that is also perhaps suggested by Figure~\ref{fig:COHAEllipses}, which shows a similar density of verbs along the branch with positive $s$ (towards irregularity) and negative $s$ (towards regularity). \cite{YangTolerancePEnglish} suggest that irregularisation may occur if the number of verbs within an irregularity class is high enough to surpass a productivity threshold. Following \cite{ref:BybeePhonology,ref:PrasadaAnalogy}, both \cite{ref:Cuskley} and \cite{ref:FITApplication} propose phonological analogy as a potential mechanism for irregularisation. Couched in the terms of the present work, this would correspond to the general rule (adding \mbox{-ed}) contributing a negative value to $s$ whilst rules that apply only to a specific subset of verbs contribute a positive value to $s$. Note that we do not necessarily imply that these contributions are additive: for example, in optimality theory \cite{ref:Optimality,ref:OptimalPhonology}, higher-ranked rules take precedence over lower-ranked rules. In general, we may regard opposing forces on linguistic variation as arising from competing motivations which have been discussed in a variety of language change contexts (e.g., \cite{ref:CompetingMotivations,ref:CompetitionModel1987,ref:CompetitionModel1989, haiman1983iconic, kirby1997competing, hawkins2004efficiency}). By whatever mechanism this opposition is resolved, an overall positive $s$ value here indicates that the irregularising rule is dominant.

In this section, we investigate a distinct motivation that may favour irregularisation, namely the phonological simplicity that is afforded by omitting a sound repetition that would occur under application of the regular rule. Specifically, we consider verbs whose infinitives end in alveolar stops (/d/ or /t/) and have an irregular past form where the regular -ed termination is omitted. Examples include \textit{I bled} instead of \textit{I bleeded} or \textit{she bet} instead of \textit{she betted}. Verbs where devoicing of final /d/ or changes in the root vowel take place on top of the omission of the termination are also considered. Thus, we hypothesise that the regular form is preferred from the point of view of inflectional simplicity (i.e. using the regular everywhere leads to a simpler inflectional system), while the irregular form is favoured by phonological simplicity. By applying the BwS algorithm to estimate the $s$ parameter (and in particular, its sign), we can assess how these competing motivations play out.

For this investigation we switch to the 2019 English Google Books corpus \cite{ref:GoogleBooks}, as the number of verbs falling into this category and whose past tense forms are both sufficiently frequent and can be reliably identified is relatively small. The larger size of Google Books relative to COHA allows more examples to be included. We identified 19 English verbs whose irregular and regular forms both show usage above 1\% at least in one 5-year bin in the Google Books corpus in the considered time frame (1809 to 2009). These verbs are: \textit{bend}, \textit{bet}, \textit{bite}, \textit{blend}, \textit{build}, \textit{fit}, \textit{glide}, \textit{knit}, \textit{light}, \textit{pat}, \textit{plead}, \textit{quit}, \textit{slide}, \textit{speed}, \textit{spit}, \textit{thrust}, \textit{tread}, \textit{wed}, and \textit{wet}. A difficulty in the analysis is that the irregular past-tense form can coincide with certain present-tense forms. A major exception is when the verb is preceded by a third-person singular pronoun (e.g., the present \textit{he bets} versus the irregular past \textit{she bet}), which can easily be distinguished in the bigram dataset. We recognise that this separation is not perfect: for example, certain English varieties do not use the third person marker -s, but we consider the effect of these contributions to be negligible in the corpus. We also kept only those cases where the pronoun was judged to appear at the start of a sentence (by virtue of capitalisation), so as to exclude contexts where the pronoun is followed by the infinitive in a question or an inversion. Again there are situations where capitalised pronouns can appear mid-sentence, but these are also rare. With this, total counts of usage for verbs with non distinct irregular past tense forms range roughly between $2,600$ (\textit{knit}) and $120,000$ (\textit{pat}), while counts for verbs whose irregular past tense is distinct from their base form range between $600,000$ (\textit{glide}) and $40,000,000$ (\textit{build}).

In order to formally test whether a potential bias towards irregularisation is significant, a similar analysis was carried out on a baseline set of 34 English verbs whose base form does not end in /d/ or /t/. Data was extracted from Google Books and all verbs satisfy the same conditions on minimal usage in the time frame of interest (1809-2009). The chosen verbs are: \textit{awake}, \textit{blow}, \textit{burn}, \textit{catch}, \textit{cleave}, \textit{creep}, \textit{dive}, \textit{dream}, \textit{dwell}, \textit{freeze}, \textit{grow}, \textit{hang}, \textit{heave}, \textit{hew}, \textit{kneel}, \textit{lean}, \textit{leap}, \textit{learn}, \textit{shake}, \textit{shear}, \textit{shine}, \textit{slay}, \textit{slink}, \textit{smell}, \textit{sneak}, \textit{spell}, \textit{spill}, \textit{spoil}, \textit{strew}, \textit{string}, \textit{strive}, \textit{swell}, \textit{wake} and \textit{weave}. Total usage for these verbs in the Google Books corpus for the specified period ranges between $211,000$ tokens (\textit{slink}) and $31,900,000$ (\textit{learn}), in the same orders of magnitude as the /d/,/t/ set.

\begin{figure}[tbp]
\begin{center}
\includegraphics[width=0.8\linewidth]{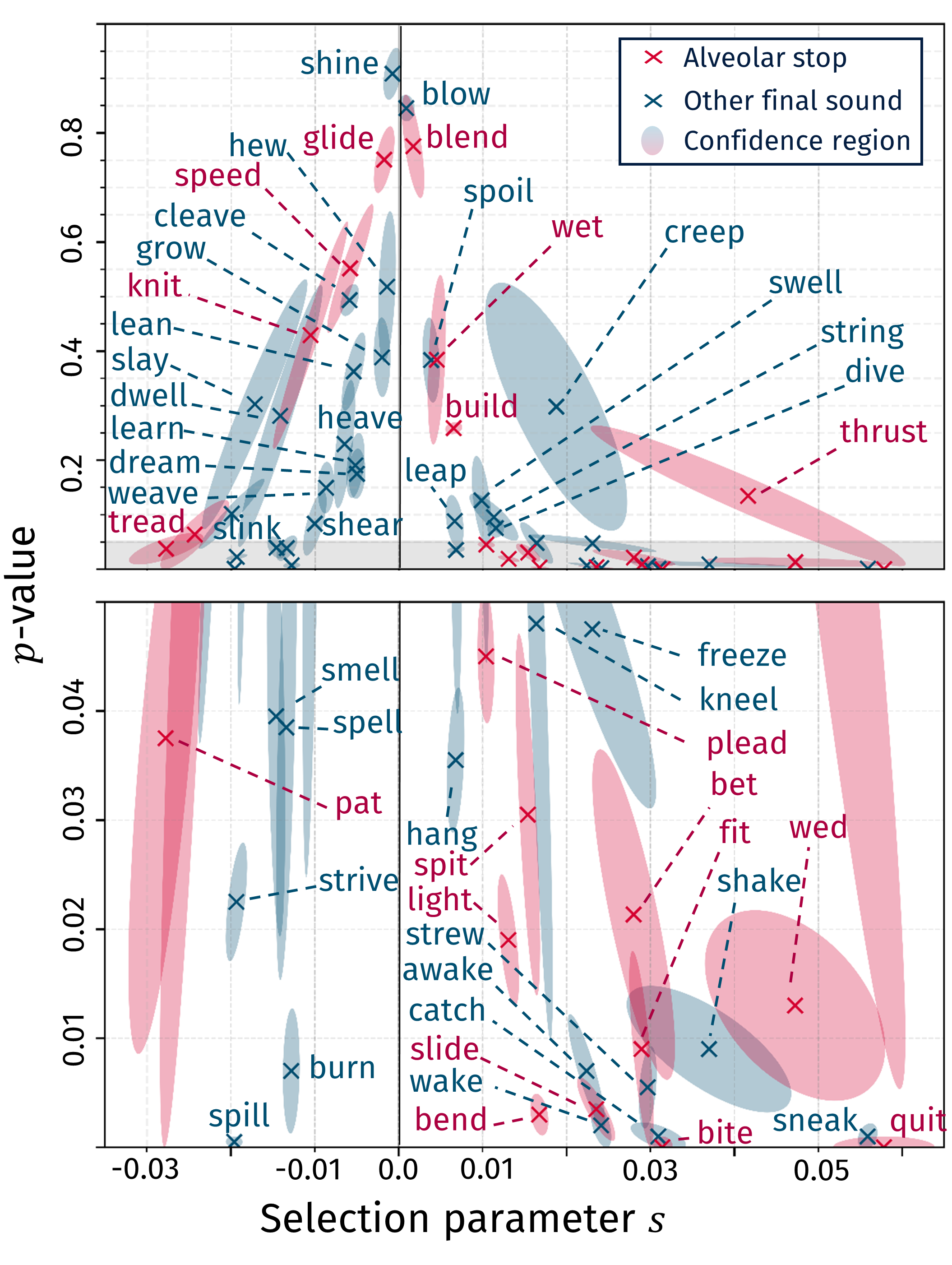}
\end{center}
\caption{Parameter estimates for verbs ending in alveolar stops (red) and verbs in the baseline set (blue) in the Google Books data set. The top panel shows the entire range of drift $p$-values and includes all 53 verbs. The bottom panel is restricted to $p<0.05$, thus focusing on verbs that are likely to be undergoing directed selection. The distribution of verbs in the alveolar stop set seems to be skewed to the region where $s>0$ and $p<0.05$, suggesting they are more likely to be irregularising than the other verbs.}
\label{fig:EnglishResults}
\end{figure}

The maximum likelihood parameters for these 53 verbs are given in Section B of the appendix. Here, we visualise our findings by plotting ellipses in the plane spanned by the selection strength and the indicator of selection, following the same procedure as previously described for the COHA verbs, albeit with the addition of a 5-year temporal binning strategy. With this, each ellipse in the $s$-$p$ plane for each verb is produced by averaging the results of the analyses of at most four temporal binnings. We recall that these ellipses characterise the variability in these parameters as the temporal binning is varied. The upper panel in Figure~\ref{fig:EnglishResults} shows the results for all 53 verbs.

For the purpose of comparing the two sets of verbs, we partition the $s$-$p$ plane into four regions: those with positive or negative selection strengths; and those where the $p$-value falls above or below $0.05$. The lower panel of Figure~\ref{fig:EnglishResults} zooms in on this latter region, which we may regard as showing evidence of selection. In both panels, red crosses and ellipses correspond to verbs ending in alveolar stops, while blue crosses and ellipses correspond to verbs in the baseline set. Given our interest in irregularisation, three groups of verbs can be identified. 16 verbs (\textit{awake}, \textit{bend}, \textit{bet}, \textit{bite}, \textit{catch}, \textit{fit}, \textit{hang}, \textit{light}, \textit{quit}, \textit{shake}, \textit{slide}, \textit{sneak}, \textit{spit}, \textit{strew}, \textit{wake}, \textit{wed}) have their confidence regions (ellipses) completely contained in the region of likely selection of the irregular form ($p<0.05$ and $s>0$, lower-right panel). Of those, 9 are in the alveolar stop set and 7 are in the baseline set. Six verbs (\textit{freeze}, \textit{kneel}, \textit{leap}, \textit{plead}, \textit{swell}, \textit{thrust}) have confidence regions only partially contained in this region of the $s$-$p$ plane, indicating that, while selective forces towards the irregular form are a plausible explanation to their dynamics, the pure drift hypothesis cannot be confidently ruled out. The remaining 31 verbs (8 in the alveolar stop set, 23 in the baseline set) have confidence regions contained entirely outside this region of likely irregularisation. 

These results suggest that verbs in the alveolar stop set are more likely to be selected towards their phonologically simpler irregular form than their counterparts in the baseline set. To test the significance of these findings, we construct the $2\times3$ contingency table shown in Table \ref{tab:GTest}, where one dimension expresses belonging to the alveolar stop or the baseline sets, while the other dimension expresses whether the verbs' ellipse falls in the irregularisation region in the bottom panels of Figure \ref{fig:EnglishResults}. The $p$-value for the null hypothesis that the baseline and alveolar stop verbs are drawn from the same distribution is $0.031$, as obtained by applying the G-test of goodness-of-fit to the contingency table \cite{McDonaldGTest}. This indicates that the specific rule favouring phonological simplicity likely outcompetes a general tendency towards regularity.

\begin{table}[tbp]
    \centering
    \begin{tabular}{c|c|c|c}
        & Irregularising & Inconclusive & Non-irregularising \\
        Alveolar stop set & 9 & 2 & 8 \\
        Baseline set & 7 & 4 & 23 \\
    \end{tabular}
    \caption{Contingency table for the comparison of irregularising behaviour between the set of verbs ending in alveolar stops and the baseline set. Irregularisation is significantly more common amongst verbs ending in alveolar stops, with a $p$-value of 0.031 as provided by the G-test.}
    \label{tab:GTest}
\end{table}

It is possible that other effects may be responsible for this subset of verbs tending to irregularise. For example, it is well understood that higher frequency items tend to tolerate greater irregularity \cite{ref:BybeeFrequency}. Given the selection criteria imposed to arrive at the set of 12 verbs in this analysis, it is possible that the sample is skewed towards higher frequency and more irregular forms. However, as noted, the total token counts for both the baseline set and the alveolar stop set span similar ranges, and also have similar averages (of around $5$ million for both sets). Therefore we consider this alternative explanation unlikely.

This is not the only phonological conditioning on irregularisation that can be inferred from Figure~\ref{fig:EnglishResults}. The subset of verbs ending in a short vowel plus a lateral (dwell, smell, spell, spill, swell) seem to be a lot more likely to regularise under selection than other verbs in the study. A similar G-test to the one performed on the alveolar stop test on Table \ref{tab:GTest} reveals that this tendency is significant with $p<0.003$. The origin of this tendency is, however, unclear.

To summarise, in this section we have shown that by focussing on a subset of verbs that are subject to specific combination of competing motivations, the Wright-Fisher model combined with the BwS approximation can be used to determine the net effect of this competition. Specifically, we have acquired evidence that phonological simplicity dominates inflectional simplicity in this competition, suggesting perhaps that this is an instance of an OCP constraint (Obligatory Contour Principle, \cite{LebenOCP,StermbergerHaplology}). OCP constraints disfavour pairs of identical or near-identical consonants from being in close proximity to each other. In particular, the constraint here appears to be an OCP-place constraint (\cite{McCarthy1986OCPplace,FrischOCPplace,pozdniakov2007OCPplace}), meaning that it does not just affect identical consonants, but all alveolar stops independently of voicing.

\subsection{Spanish spelling reforms}

So far we have assumed that the evolutionary parameters (the effective population size $N$ and the selection strength $s$) have been constant over time. In the case of competition between regular and irregular verbs this is a reasonable assumption, due to the factors favouring one over the other likely being cognitive or linguistic in origin. By contrast, social pressures like prestige, taboo, or language contact \cite{ref:HandbookHistoricalSociolinguistics,ref:LabovLinguisticChangeSocial,ref:UnderstandingLanguageChange} are inherently time-dependent, and we may expect the selection strength in particular to change over time. Here, we investigate this possibility in the context of a purposeful change made by a regulating institution through prescriptive grammar and spelling rules \cite{ref:NormativeLanguageChange,ref:LanguagePlanning}, the acceptance or rejection of which we expect to be reflected by a change in the value of $s$. 
While well established algorithms like change-point analysis \cite{taylor2000changepoint} exist for the detection of change in time series, these suffer from shortcomings that make them inadequate for a more nuanced analysis of change in language and culture. First, change-point analysis is based on the assumption that the data is distributed around constant average before and after a change, which changes the value of said average instantaneously. This makes this methodology only fit for the detection of rapidly occurring S-shaped curves of language change, where the usage frequency of a variant quickly changes and stabilises. Secondly, change-point analysis provides no extra linguistic information, as it does not assume a model of the underlying evolutionary dynamics.
\cite{ref:Us} solve this issue by setting out a procedure for estimating times at which the parameters $N$ and $s$ change, thus measuring changes in the evolutionary dynamics of the data rather than its average. We briefly recapitulate and then apply this method below.

The specific changes of interest are spelling reforms in Spanish that were introduced by the Real Academia Espa\~nola (RAE), the central regulatory institution of the standard Spanish language. Since its creation in 1713, the RAE has regulated Spanish orthography following the phonemic principle over etymological or conservative approaches \cite{ref:OrthographiesEurope}. We study words affected by one of the following reforms: (A) The simplification of the \textless ss\textgreater\, digraph to a single \textless s\textgreater\, in 1763, due to the different sounds that both spellings represented having merged in the 16th century \cite{ref:RAE1763}; (B) The replacement in 1815 of etymological \textless x\textgreater\, with \textless j\textgreater\, in all non word-final contexts where it represented the phoneme /x/ \cite{ref:RAE1815}; (C) The replacement, also in 1815, of \textless y\textgreater\, with \textless i\textgreater\, in all non word-final closing diphthongs; (D) The reversal of accentuation rules for words ending in \textless n\textgreater, introduced in 1881. This reform stipulated that words ending in \textless n\textgreater\, with a tonic last syllable had to be accentuated, while words ending in \textless n\textgreater\, with a tonic penultimate syllable lost their previously prescribed accent \cite{ref:RAE1881}. We treat words that gain an accent and words that lose an accent as independent sets (D.1 and D.2, respectively). 

We now seek to estimate the time at which each reform occurred by appealing only to the time series data and no external information. The basic idea (see also \cite{ref:Us}) is to allow different parameter combinations $(N,s)$ to apply before and after a time $T$. That is, for $t<T$ the Wright-Fisher model with parameters $(N_1,s_1)$ applies, and for $t>T$ the parameters $(N_2,s_2)$ apply. The data likelihood, obtained by combining Eqs.~(\ref{eq:Likelihood}) and (\ref{eq:BWS}), is then maximised with respect to all five parameters (i.e., $N$ and $s$ each before and after the change, and the time $T$ of the change itself).

After identifying the time $T$ that maximises the data likelihood, one needs to determine if the additional complexity of the five-parameter model is compensated by a sufficiently improved description of the data. To achieve this, we obtain an empirical $p$-value for the null hypothesis that the selection strength $s$ was constant over the entire time period by following a procedure similar to that described in Section~\ref{sec:Method1}. Specifically, we determine the maximum likelihood values of $N$ and $s$ without a change point, and generate 500 synthetic time series that match the length of the observed series with these parameter values. For each of these time series, we then optimise the five-parameter likelihood that applies when the selection strength changes at a single point in time. An empirical $p$-value is then given by the fraction of such time series whose five-parameter likelihood exceeds that of the real trajectory. Although computational constraints limit the number of synthetic time series that can be analysed this way, we find that situations where the five-parameter fit has a high likelihood are extremely rare, and there is little to be gained by estimating their rarity to greater precision. One can then apply a threshold, e.g., $p<0.05$, to decide whether to accept the more complex model. Having split the time series once, one can apply the method again to each sub-series, thereby identifying secondary change points. This procedure terminates when none of the sub-series admits a subdivision that yields a sufficiently improved description of the data according to the threshold that has been imposed.

To apply this method to the Spanish Spelling reforms, we identify a set of commonly used words that are affected by each one, and average the relative frequencies of usage of their old spellings over all members of each set. The number of words in each set ranges from 16 to 27. The exact sets are specified in Section C of the appendix. This procedure generates a single effective time series for each of the reforms, and has been found effective in related corpus analyses \cite{ref:DynamicsofNormChangeCulturalApplicationAmato}.

While this averaging over sets of words decreases the sampling noise in the data and increases the inferential power of the analysis, cultural data still suffers from issues that may affect the applicability of the method. Particularly, corpora tend to contain lower token counts in earlier time periods. When translated to frequency time series, this leads to greater sampling noise fluctuations that may be misidentified as changes in the effective population size parameter $N$. This issue can be remedied by applying a sampling error equalisation algorithm, as laid out by \cite{ref:Us}. This method creates subsamples of the larger token counts in the data set, in such a way that sampling effects are of equal magnitude throughout the data. In this way, any significant changes in $N$ detected by the method must be due to changes in effective population size parameter, and not a consequence of unequal sampling noise.

\begin{figure}[tbp]
\begin{center}
\scalebox{0.35}{\includegraphics{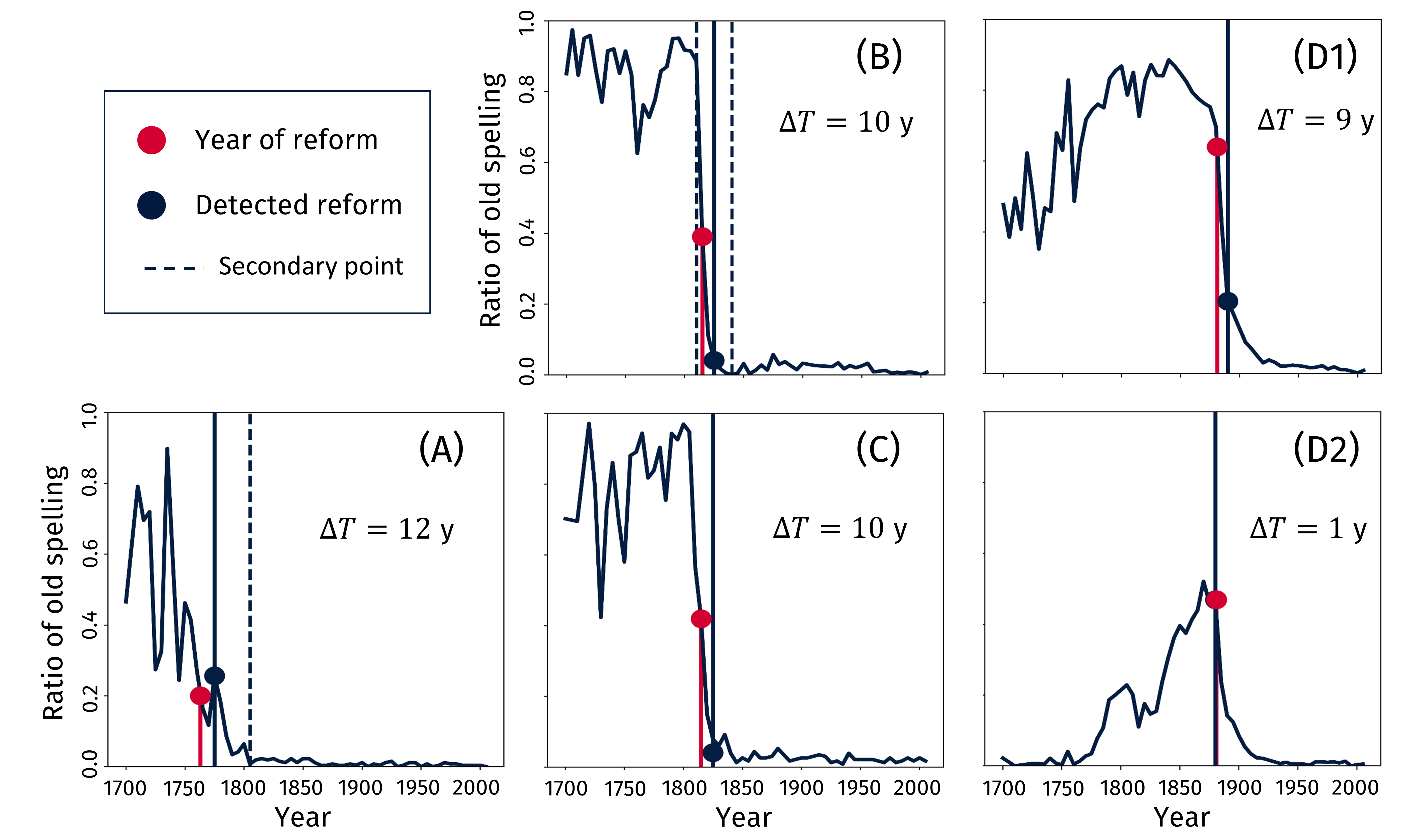}}
\end{center}
\caption{Application of the BwS algorithm for the detection of changing forces to the reference data set of Spanish spelling reforms in the 2019 Spanish 1-gram Google Books corpus, with temporal binning of the frequency data of 5 years. For each set of words that undergo a rule change, the ratio of usage of the old form is plotted over time.  The ratio of usage of all old forms converges to zero after each reform.  Red dots with solid vertical lines represent the year of publication of the RAE spelling reforms \protect\cite{ref:RAE1763,ref:RAE1815,ref:RAE1881}. Dark blue dots with solid vertical lines represent the year at which selection strengths changed as detected by the maximum likelihood method with a $p$-value below $0.05$. These fall within a period $\Delta T$ of 12 years or less relative to the date of the reform. Note that the temporal resolution of the time series is of 5 years, so an error of 10 years is equivalent to just two data points. Dashed vertical lines represent secondary points of change in evolutionary parameters, also detected with a $p$-value below 0.05. The number of such secondary points depends on the time series.}
\label{fig:SpanishResults}
\end{figure}

Our results are shown in Figure~\ref{fig:SpanishResults}. Despite the aggregation of words within each category (to improve the inferential power) and 5-year bins (to reduce computational effort), we find that the resulting trajectories are still subject to considerable fluctuations. The frequency plotted is that of the deprecated variant, which we find is eliminated in all five cases---this highlights the acceptance and influence of the Real Academia Espa\~{n}ola amongst the literate population. We show with a red line and dot the time at which the reform was introduced, and with a black dot and solid line the first time $T$ at which subdividing the time-series improves the fit to the data, with a $p$-value threshold of $p=0.05$ applied.

\begin{table}[tbp]
    \centering
    \begin{tabular}{rc|c|c|c}
        &Reform & Detected year & $s$ before division & $s$ after division  \\
        (A)& \textless ss\textgreater to \textless s\textgreater & 1775 & -0.008 & -0.37 \\
        (B)& \textless x\textgreater to \textless j\textgreater & 1825 & -0.29 & -0.07 \\
        (C)& \textless y\textgreater to \textless i\textgreater & 1815 & -0.30 & -0.06 \\
        (D1)& Accentuation & 1890 & -0.05 & -0.21 \\
        (D2)& Loss of accent & 1880 & 0.13 & -0.53 \\
    \end{tabular}
    \caption{Maximum likelihood estimates of the first detected time at which the selection strength changed, and its values before and after the change, for the five Spanish reform categories. All changes significant with $p<0.002$.}
    \label{tab:SValuesSpanish}
\end{table}

In all five cases, we find evidence that the selection $s$ changed significantly over time. In each case, the first detected change point falls within twelve years of the reform being introduced, even when the trajectory is strongly fluctuating. We note that the algorithm does have a tendency to detect the reform after it has occurred, rather than at its inception. This is due to the algorithm not distinguishing past from present, making both the beginning and the end of the sharp decline following a reform are considered equivalent.

Further iterating the algorithm, we can further subdivide the time series, as described above. In doing so, we detect secondary time divisions (dashed lines in Figure~\ref{fig:SpanishResults}), whose $p$-values are below $0.05$. In time series (B), the earlier secondary point detects the beginning of the rapid decline in usage that was deemed less significant than the end by the first application of the algorithm. The later secondary point in (B) and the secondary point in (A) are not associated to documented reforms, and may reflect slight changes in social attitudes or simply be quirks of the data.


Table~\ref{tab:SValuesSpanish} further records the $s$-values before and after the main change point. All $s$, $N$ and $p$-values for every main and secondary point detected by the algorithm can be found in Section D of the appendix. For categories (A), (D.1) and (D.2), we find that the $s$ value decreases after the detected year of the reform, corresponding to an acceptance of the reform by the speech community. The other two categories however show the opposite trend, with the $s$ value becoming less negative across the reform. We note from Figure~\ref{fig:SpanishResults} that both categories (B) and (C) feature a rapid elimination of the deprecated form, and that this change was in progress before the reform was introduced. It has been suggested that in many cases, language reforms tend to reflect pre-existing trends, as opposed to actuating the change \cite{ref:FromAbove}. Our analysis provides further evidence of this, and further suggests that the impact of the reform on the speech community may be limited in such cases.

In summary, this analysis indicates that the BwS method can be used successfully to characterise evolutionary forces that change over time from time series data alone. As an unsupervised method, it does not rely on any prior knowledge as to when the change may have occurred, although it does benefit from a large sample size being available, obtained here by aggregating multiple instances of a change together. We have found that the estimated time at which the selection strength changed corresponds well with the time at which the corresponding reform was introduced, and comparing these strengths before and after the reform allows us to assess its impact on the speech community.

\section{Discussion}\label{sec:disco}

In this work we have applied an algorithm for the quantitative study of evolutionary time series \cite{ref:Us} to instances of competition in language change. This algorithm is based on likelihood-maximisation methods and the Beta-with-Spikes (BwS) approximation to the Wright-Fisher model. The applicability of the Wright-Fisher model was justified through both theoretical considerations \cite{ref:Hull,ref:CroftLanguageChange} and its manifestation as an agent-based model of language change from various starting points \cite{ref:UtteranceSelection,ref:WordsAsAlleles,ref:RichardPLOS}.

In Section~\ref{sec:bws}, we demonstrated that the BwS method better captured the statistical properties of the Wright-Fisher model than the normal approximation that has been used elsewhere \cite{ref:FITApplication}. In particular, it deals better with situations where variant frequencies are close to 0 or 1, which arises in the case when selection serves to eliminate linguistic variation across the speech community. Through refinements to the original BwS method of \cite{ref:BWS1} that are detailed by \cite{ref:Us}, we further gain accuracy in regimes where the selection strength is large. 

Our first application was to the set of 36 COHA verbs previously investigated by other methods \cite{ref:FITApplication,ref:ProblemsofFIT,ref:NeuralNetworks}. In particular, we found that even when the Frequency Increment Test (FIT,  \cite{ref:FITApplication,ref:ProblemsofFIT}) delivered unreliable results due to shortcomings of the normal approximation that it relies on, we obtained evidence of selection that was broadly consistent with that obtained within a Time Series Classification (TSC, \cite{ref:NeuralNetworks}) which took the complementary approach of training neural networks with artificial time series. The present method further delivered graded measure of the extent to which the historical changes are consistent with drift (in the form of a null hypothesis $p$-value) along with maximum likelihood estimates of parameters in the Wright-Fisher model.

A degree of care is needed when interpreting this $p$-value. All evolutionary trajectories are likely to be the product of some combination of drift and selection. The key question is whether their respective contributions can be distinguished. For example, a variant could be strongly selected for (large $s$) but subject to sufficiently large fluctuations (small $N$) that the systematic effects of selection are masked. The $p$-value is therefore a measure of the extent to which fluctuations alone could account for the changes that have been observed. If one chooses to apply the conventional significance threshold for rejecting this null hypothesis ($p<0.05$), we find consistency with \cite{ref:FITApplication}'s observation that the evolution of many verbs appears to be dominated by drift.

A second important question is whether these fluctuations are a consequence of the finite number of tokens available for analysis in historical corpora, or an intrinsic property of the language dynamics within the speech community. One way to gain an insights into this question is to compare results obtained with different temporal binnings  (Figures~\ref{fig:COHAResults} and \ref{fig:COHAEllipses}), since wider bins contain more points and should reduce fluctuations due to sampling. If sampling effects were dominating, we would expect to see the $p$-value for the drift hypothesis to decrease as the bins are widened (i.e., increasing darkness in Figure~\ref{fig:COHAResults}). This happens for some, but not all the verbs in the intermediate region, suggesting that drift may be the dominant factor in the evolution of a substantial fraction of the COHA verbs (again, consistent with \cite{ref:FITApplication}). A more rigorous answer to this question could be obtained by incorporating finite sample-size effects into the data likelihood function used in the analysis. This is however likely to be computationally demanding, and we leave this possibility for future work.

In this work, we found plotting ellipses that indicate the variation in estimates of the selection strength and the $p$-value for the drift hypothesis helpful to understand which variants are more likely to have been selected for. A comparison between a baseline set of verbs from the Google Books corpus and a set where the past tense is formed by deletion of a repeated consonant reveals that they are distributed differently across the space of selection strengths $s$ and drift $p$-values. Specifically, we found that the phonological simplicity arising from coalescence or omission of the /\textipa{Id}/ termination tended to be favoured over the inflectional simplicity of the regular form. In principle, the method we have set out here could be used to determine the relative importance of other pairs of constraints that correspond to opposing selective forces.

Finally, we showed that the method could be applied also to changes that do not have a cognitive origin and manifest as the selection strength changing over time. We studied the dynamics of word spellings in Spanish before and after reforms introduced by their central regulatory institution, the Real Academia Espa\~{n}ola. We found that each of the changes was much better described by a model in which the selection strength changed at one or more points in time, and that the primary change point corresponded well with that at which the reform was introduced. This is despite the presence of noise on the time series data. Since changes in selection strength could derive from a variety of social and cultural factors, and indeed apply to cultural evolutionary processes beyond language, this method for automated detection of societal trends and shifts could have broad applicability.


Despite these promising results, there are inevitably some limitations. Chief among these is an inability to separate different contributions to the selective pressures acting on the system. Therefore, although it is possible to use this data to determine that selection has favoured one variant over another over time, and to estimate the strength of the effect, we have had to appeal to additional information to relate to likely causes of selection. This, however, is a problem intrinsic to the Wright-Fisher model with selection and not specific to the BwS method: the Wright-Fisher model contains only a single parameter $s$ that characterises all systematic contributions to changes in variant frequencies. 

This oversimplification of the contributing factors to language change stems, at least in part, from the underlying assumption that the competition between forms (e.g. irregular and regular verbal forms) occurs in isolation, uninfluenced by the competition dynamics of related forms (e.g. irregular and regular forms of other verbs). \cite{YehCulturalLinkage} and \cite{BuskellSystemsApproach} argue, in the context of cultural evolution, that cultural change may arise as an emergent phenomenon when cultural traits are interconnected. It is possible, then, that emergent system-level effects may account for significant changes in usage frequency of variants that are not favoured individually by any social or inductive bias. More refined models, ones that account for the complex web of interconnected forms and functions present in language, may be able to differentiate between these systemic effects and those affecting individual variants. Such models might allow more information to be extracted from corpora without the need for additional information.

Nevertheless, we have shown that it is possible to draw inferences about contributions to selection from different sources (as was done in the analysis of competition between regular and irregular forms in English verbs) and quantify the impact of social factors (as was done in the language reform example). By appealing to a wider range of corpora and instances of change, it may become possible to identify general mechanisms that are invariant over time and operate cross-linguistically, and are thus informative about language universals in general. Furthermore, the method is not specific to linguistic variation, and could be used to address similar questions in other instances of cultural evolution.

\section*{Acknowledgements}

We thank Johanna Basnak (University of Edinburgh) for helpful discussions.

\section*{Data availability}
\label{sec:DataAvailability}

The code and data are available here: \href{https://osf.io/qxgnj/?view_only=de983f003307492bb6dd777ee3e36a39}{link}

\section*{Funding information}

Juan Guerrero Montero holds a Principal's Career Development scholarship awarded by the University of Edinburgh.

\section*{Competing interests}

The authors declare no competing interests.

\printbibliography 

\newpage
\appendix

\section{Maximum likelihood parameters for the COHA verbs}
\label{app:coha}

In the following tables we quote the maximum-likelihood estimates of the parameters in the Wright-Fisher model obtained by applying the Beta-with-Spikes method outlined in the main text to frequency counts derived from the COHA corpus. Each table corresponds to a different binning strategy: for example, in the first table, frequency counts from each period of 10 consecutive years are aggregated to form a single frequency estimate for the corresponding time period. 

Two different effective population sizes $N$ are quoted: one (`for drift') under the assumption that $s=0$, and the other (`for selection') that is obtained when both $N$ and $s$ are optimised via the maximum likelihood analysis. The $p$-value is the empirical $p$-value for the drift hypothesis, obtained as described in the Section \ref{sec:methods} of the main text. The maximum likelihood values are all quoted to three significant figures, and the $p$-values to two significant figures.

\clearpage

\subsection*{10-year bins}

\begin{center}
    \begin{tabular}{|c|c|c|c|c|}
        Verb & $N$ for drift & $N$ for selection & $s$ & $p$-value \\
        awake & 1820 & 1990 & 0.021 & 0.11\\ 
        build & 2820 & 3130 & 0.026 & 0\\ 
        burn & 410 & 542 & -0.05 & 0\\ 
        catch & 3690 & 4170 & 0.031 & 0.028\\ 
        dive & 145 & 147 & 0.0092 & 0.43\\ 
        draw & 2880 & 2880 & 0.00067 & 0.96\\ 
        dream & 219 & 233 & -0.036 & 0.002\\ 
        dwell & 568 & 554 & -0.029 & 0.002\\ 
        grow & 4510 & 4770 & 0.031 & 0.032\\ 
        hang & 1520 & 1830 & 0.048 & 0\\ 
        hear & 6160 & 6920 & 0.047 & 0.044\\ 
        heave & 147 & 145 & 0.0069 & 0.68\\ 
        kneel & 360 & 362 & 0.0028 & 0.82\\ 
        knit & 121 & 122 & -0.0043 & 0.77\\ 
        know & 4240 & 4350 & 0.0035 & 0.65\\ 
        lay & 4070 & 4250 & 0.002 & 0.47\\ 
        lean & 753 & 926 & -0.053 & 0\\ 
        leap & 313 & 346 & 0.022 & 0.16\\ 
        learn & 652 & 847 & -0.052 & 0\\ 
        light & 294 & 368 & 0.021 & 0.03\\ 
        plead & 1500 & 1590 & -0.012 & 0.19\\ 
        quit & 790 & 927 & 0.022 & 0\\ 
        shine & 1330 & 1320 & -0.014 & 0.17\\ 
        smell & 319 & 399 & -0.036 & 0.004\\ 
        sneak & 415 & 453 & 0.023 & 0.068\\ 
        speed & 299 & 327 & 0.024 & 0.052\\ 
        spell & 288 & 307 & -0.017 & 0.31\\ 
        spill & 304 & 357 & -0.032 & 0\\ 
        spoil & 223 & 226 & -0.0036 & 0.79\\ 
        strew & 127 & 127 & -0.0023 & 0.86\\ 
        tell & 3760 & 3960 & 0.018 & 0.37\\ 
        throw & 2090 & 2180 & 0.012 & 0.25\\ 
        wake & 815 & 928 & 0.015 & 0.012\\ 
        weave & 460 & 457 & -0.007 & 0.45\\ 
        wed & 116 & 120 & 0.026 & 0.03\\ 
        wet & 201 & 201 & 0.0024 & 0.88\\
    \end{tabular}
\end{center}

\subsection*{20-year bins}

\begin{center}
    \begin{tabular}{|c|c|c|c|c|}
        Verb & $N$ for drift & $N$ for selection & $s$ & $p$-value \\
        awake & 5130 & 5730 & 0.011 & 0.13\\ 
        build & 4850 & 5290 & 0.0064 & 0.27\\ 
        burn & 615 & 909 & -0.042 & 0\\ 
        catch & 9890 & 12000 & 0.02 & 0.058\\ 
        dive & 483 & 546 & 0.0095 & 0.24\\ 
        draw & 6690 & 6660 & 0.0028 & 0.84\\ 
        dream & 296 & 326 & -0.034 & 0.004\\ 
        dwell & 1080 & 1430 & -0.04 & 0\\ 
        grow & 17600 & 17600 & -4.2e-05 & 1.0\\ 
        hang & 4260 & 4310 & 0.0052 & 0.58\\ 
        hear & 16400 & 17500 & 0.013 & 0.39\\ 
        heave & 353 & 350 & 0.0018 & 0.85\\ 
        kneel & 1580 & 1640 & 0.0018 & 0.76\\ 
        knit & 244 & 251 & -0.0049 & 0.66\\ 
        know & 8890 & 8900 & 4.1e-05 & 1.0\\ 
        lay & 9110 & 10200 & 0.0016 & 0.43\\ 
        lean & 1060 & 1990 & -0.063 & 0\\ 
        leap & 498 & 616 & 0.019 & 0.1\\ 
        learn & 937 & 1910 & -0.054 & 0\\ 
        light & 360 & 834 & 0.022 & 0.006\\ 
        plead & 2100 & 2110 & -0.00072 & 0.94\\ 
        quit & 1140 & 1690 & 0.025 & 0\\ 
        shine & 2500 & 2740 & -0.018 & 0.066\\ 
        smell & 500 & 1440 & -0.036 & 0\\ 
        sneak & 674 & 871 & 0.024 & 0.02\\ 
        speed & 807 & 941 & 0.014 & 0.14\\ 
        spell & 498 & 1310 & -0.042 & 0\\ 
        spill & 688 & 1220 & -0.025 & 0\\ 
        spoil & 565 & 747 & -0.028 & 0.018\\ 
        strew & 334 & 334 & 0.0012 & 0.87\\ 
        tell & 7620 & 7570 & 0.012 & 0.62\\ 
        throw & 9380 & 9420 & 0.00046 & 0.94\\ 
        wake & 1160 & 1650 & 0.015 & 0.014\\ 
        weave & 1020 & 1000 & -0.011 & 0.14\\ 
        wed & 147 & 216 & 0.028 & 0.1\\ 
        wet & 1100 & 1140 & 0.0021 & 0.78 \\
    \end{tabular}
\end{center}

\subsection*{40-year bins}

\begin{center}
    \begin{tabular}{|c|c|c|c|c|}
        Verb & $N$ for drift & $N$ for selection & $s$ & $p$-value \\
        awake & 9510 & 20300 & 0.016 & 0.042\\ 
        build & 5780 & 8090 & 0.0098 & 0.16\\ 
        burn & 1770 & 15100 & -0.017 & 0.014\\ 
        catch & 15000 & 40800 & 0.027 & 0.002\\ 
        dive & 697 & 879 & 0.0072 & 0.49\\ 
        draw & 30800 & 48600 & 0.0057 & 0.52\\ 
        dream & 824 & 868 & -0.0075 & 0.52\\ 
        dwell & 1430 & 2060 & -0.037 & 0\\ 
        grow & 31200 & 31100 & -0.0012 & 0.95\\ 
        hang & 146000 & 148000 & -0.00099 & 0.68\\ 
        hear & 20100 & 20300 & 0.0073 & 0.68\\ 
        heave & 743 & 835 & -0.0046 & 0.66\\ 
        kneel & 3090 & 6490 & 0.0062 & 0.25\\ 
        knit & 252 & 261 & -0.0027 & 0.83\\ 
        know & 8310 & 8720 & 0.0025 & 0.75\\ 
        lay & 9020 & 13400 & 0.0021 & 0.35\\ 
        lean & 2810 & 5530 & -0.025 & 0.008\\ 
        leap & 775 & 968 & 0.015 & 0.14\\ 
        learn & 2680 & 9430 & -0.023 & 0.006\\ 
        light & 318 & 1110 & 0.019 & 0.03\\ 
        plead & 3770 & 3690 & 0.0047 & 0.62\\ 
        quit & 659 & 1610 & 0.026 & 0\\ 
        shine & 3910 & 8610 & -0.019 & 0.01\\ 
        smell & 477 & 2600 & -0.034 & 0\\ 
        sneak & 924 & 920 & 0.014 & 0.24\\ 
        speed & 1050 & 1040 & -0.00025 & 1.0\\ 
        spell & 509 & 2180 & -0.039 & 0\\ 
        spill & 712 & 2060 & -0.026 & 0\\ 
        spoil & 1350 & 2490 & -0.014 & 0.086\\ 
        strew & 569 & 623 & 0.0043 & 0.65\\ 
        tell & 11000 & 11700 & 0.025 & 0.46\\ 
        throw & 6510 & 6630 & -0.0017 & 0.93\\ 
        wake & 908 & 2810 & 0.019 & 0.014\\ 
        weave & 1160 & 1690 & -0.017 & 0.052\\ 
        wed & 248 & 364 & 0.016 & 0.14\\ 
        wet & 2570 & 3220 & 0.0037 & 0.13\\
    \end{tabular}
\end{center}

\newpage

\section{Maximum likelihood parameters for verbs in the study of competing motivations} 
\label{app:coal}

In this appendix, we provide the corresponding tables for the set of verbs ending in alveolar stops from drawn from the Google Books corpus. Dashes mean that the corresponding time series did not have enough data points per time bin in the corresponding binning for it to be included in the study.

\subsection*{5-year bins}

\begin{center}
    \begin{tabular}{|c|c|c|c|c|}
        Verb & $N$ for drift & $N$ for selection & $s$ & $p$-value \\
        bend & 7450 & 9980 & 0.019 & 0.004\\ 
        bet & - & - & - & - \\
        bite & 4600 & 6960 & 0.029 & 0\\ 
        blend & 5110 & 5040 & 0.0037 & 0.56\\ 
        build & 21000 & 22200 & 0.0058 & 0.28\\
        fit & 586 & 596 & 0.026 & 0.036\\ 
        glide & 6520 & 6520 & -0.0026 & 0.84\\ 
        knit & - & - & - & -\\
        light & 823 & 908 & 0.011 & 0.036\\ 
        pat & 1420 & 1420 & -0.02 & 0.12\\ 
        plead & 5040 & 5160 & 0.01 & 0.054\\ 
        quit & 376 & 377 & 0.076 & 0\\ 
        slide & 2050 & 2210 & 0.023 & 0.002\\ 
        speed & 628 & 626 & -0.003 & 0.61\\ 
        spit & 757 & 847 & 0.014 & 0.064\\ 
        thrust & 2510 & 2880 & 0.074 & 0.002\\
        tread & 2980 & 3000 & -0.014 & 0.25\\  
        wed & 93.6 & 83.6 & 0.079 & 0\\
        wet & - & - & - & -\\
    \end{tabular}
\end{center}

\begin{center}
    \begin{tabular}{|c|c|c|c|c|}
        Verb & $N$ for drift & $N$ for selection & $s$ & $p$-value \\
        awake & 1390 & 2230 & 0.025 & 0\\ 
        blow & 11100 & 11200 & 0.0012 & 0.8\\ 
        burn & 1110 & 1350 & -0.012 & 0\\ 
        catch & 5070 & 6640 & 0.039 & 0\\ 
        cleave & 442 & 447 & -0.0042 & 0.54\\ 
        creep & 3820 & 4190 & 0.042 & 0.014\\ 
        dive & 757 & 766 & 0.011 & 0.088\\ 
        dream & 1830 & 1900 & -0.0041 & 0.36\\ 
        dwell & 1300 & 1300 & 0 & 1.0\\ 
        freeze & 2290 & 2510 & 0.045 & 0.008\\ 
        grow & 95400 & 95300 & -0.0012 & 0.5\\ 
        hang & 2070 & 2350 & 0.0077 & 0.052\\ 
        heave & 507 & 516 & -0.0042 & 0.58\\ 
        hew & 1320 & 1320 & -0.00037 & 0.93\\ 
        kneel & 986 & 1390 & 0.02 & 0.002\\ 
        lean & 1350 & 1380 & -0.0037 & 0.47\\ 
        leap & 1460 & 1500 & 0.0056 & 0.22\\ 
        learn & 2250 & 2360 & -0.0047 & 0.3\\ 
        shake & 3910 & 4520 & 0.065 & 0\\ 
        shear & 545 & 546 & -0.0073 & 0.26\\ 
        shine & 1380 & 1380 & 0.00025 & 0.98\\ 
        slay & 8430 & 8430 & -0.00048 & 0.98\\ 
        slink & 1310 & 1320 & -0.011 & 0.28\\ 
        smell & 710 & 799 & -0.013 & 0.018\\ 
        sneak & 2050 & 3130 & 0.055 & 0\\ 
        spell & 542 & 584 & -0.011 & 0.1\\ 
        spill & 863 & 1260 & -0.02 & 0.002\\ 
        spoil & 1370 & 1380 & 0.0046 & 0.26\\ 
        strew & 646 & 1010 & 0.028 & 0\\ 
        string & 2180 & 2650 & 0.019 & 0.038\\ 
        strive & 3520 & 3830 & -0.017 & 0.026\\ 
        swell & 1070 & 1240 & 0.011 & 0.01\\ 
        wake & 775 & 1290 & 0.025 & 0\\ 
        weave & 994 & 988 & -0.0086 & 0.16 \\
    \end{tabular}
\end{center}

\subsection*{10-year bins}

\begin{center}
    \begin{tabular}{|c|c|c|c|c|}
        Verb & $N$ for drift & $N$ for selection & $s$ & $p$-value \\
        bend & 12400 & 28100 & 0.017 & 0\\ 
        bet & 491 & 685 & 0.039 & 0\\ 
        bite & 5330 & 16600 & 0.03 & 0\\ 
        blend & 8290 & 8170 & 0.0024 & 0.68\\ 
        build & 23900 & 27000 & 0.0057 & 0.23\\ 
        fit & 1340 & 1750 & 0.03 & 0\\ 
        glide & 26300 & 25900 & 0.0014 & 0.85\\ 
        knit & - & - & - & - \\
        light & 840 & 1120 & 0.012 & 0.018\\ 
        pat & 2680 & 3090 & -0.022 & 0.028\\
        plead & 6390 & 7030 & 0.011 & 0.028\\ 
        quit & 522 & 670 & 0.052 & 0\\ 
        slide & 3240 & 3240 & 0.018 & 0.012\\ 
        speed & 444 & 445 & -0.0024 & 0.75\\ 
        spit & 1210 & 1590 & 0.013 & 0.054\\ 
        thrust & 3760 & 4630 & 0.051 & 0.006\\ 
        tread & 7190 & 8840 & -0.023 & 0\\  
        wed & 159 & 163 & 0.038 & 0.038\\
        wet & - & - & - & - \\
    \end{tabular}
\end{center}

\begin{center}
    \begin{tabular}{|c|c|c|c|c|}
        Verb & $N$ for drift & $N$ for selection & $s$ & $p$-value \\
        awake & 1310 & 3710 & 0.025 & 0\\ 
        blow & 24500 & 24700 & 0.00079 & 0.81\\ 
        burn & 1260 & 2260 & -0.013 & 0\\ 
        catch & 9740 & 28500 & 0.033 & 0\\ 
        cleave & 461 & 472 & -0.0042 & 0.51\\ 
        creep & 12600 & 14200 & 0.017 & 0.066\\ 
        dive & 1050 & 1140 & 0.012 & 0.048\\ 
        dream & 2650 & 3100 & -0.0053 & 0.16\\ 
        dwell & 1790 & 1970 & -0.014 & 0.096\\ 
        freeze & 6130 & 7090 & 0.02 & 0.038\\ 
        grow & 184000 & 187000 & -0.0018 & 0.19\\ 
        hang & 3010 & 4150 & 0.0071 & 0.04\\ 
        heave & 1130 & 1320 & -0.007 & 0.14\\ 
        hew & 7420 & 7640 & -0.0015 & 0.43\\ 
        kneel & 895 & 1560 & 0.019 & 0.008\\ 
        lean & 1350 & 1420 & -0.0038 & 0.49\\ 
        leap & 1660 & 1830 & 0.0066 & 0.1\\ 
        learn & 3130 & 3660 & -0.0054 & 0.17\\ 
        shake & 9620 & 11900 & 0.037 & 0.004\\ 
        shear & 983 & 1100 & -0.0098 & 0.06\\ 
        shine & 3570 & 3550 & -0.00087 & 0.9\\ 
        slay & 17200 & 20000 & -0.015 & 0.21\\ 
        slink & 2120 & 2230 & -0.015 & 0.12\\ 
        smell & 607 & 785 & -0.014 & 0.022\\ 
        sneak & 2190 & 3740 & 0.055 & 0\\ 
        spell & 998 & 1630 & -0.014 & 0.004\\ 
        spill & 789 & 1890 & -0.02 & 0\\ 
        spoil & 1710 & 1730 & 0.0039 & 0.26\\ 
        strew & 453 & 1130 & 0.03 & 0\\ 
        string & 8410 & 12200 & 0.012 & 0.028\\ 
        strive & 2980 & 3440 & -0.018 & 0.034\\ 
        swell & 844 & 1070 & 0.011 & 0.03\\ 
        wake & 631 & 1710 & 0.025 & 0\\ 
        weave & 2110 & 2280 & -0.0093 & 0.044 \\
    \end{tabular}
\end{center}

\subsection*{20-year bins}

\begin{center}
    \begin{tabular}{|c|c|c|c|c|}
        Verb & $N$ for drift & $N$ for selection & $s$ & $p$-value \\
        bend & 13800 & 126000 & 0.016 & 0\\ 
        bet & 677 & 913 & 0.02 & 0.056\\ 
        bite & 5670 & 18200 & 0.033 & 0\\ 
        blend & 5580 & 5580 & 0.0013 & 0.88\\ 
        build & 22600 & 28500 & 0.0064 & 0.26\\ 
        fit & 1390 & 2490 & 0.032 & 0\\ 
        glide & 102000 & 116000 & -0.0035 & 0.58\\ 
        knit & 1460 & 1480 & -0.0043 & 0.72\\ 
        light & 746 & 1850 & 0.014 & 0.008\\ 
        pat & 2660 & 6670 & -0.039 & 0\\ 
        plead & 5860 & 7540 & 0.011 & 0.044\\ 
        quit & 590 & 1020 & 0.054 & 0\\ 
        slide & 5160 & 23200 & 0.026 & 0\\ 
        speed & 434 & 430 & -0.0042 & 0.63\\ 
        spit & 1570 & 9000 & 0.017 & 0\\ 
        thrust & 4700 & 4250 & -0.013 & 0.52\\ 
        tread & 8800 & 21600 & -0.028 & 0\\ 
        wed & 310 & 353 & 0.037 & 0.012\\ 
        wet & 800 & 898 & 0.0051 & 0.6 \\
    \end{tabular}
\end{center}

\begin{center}
    \begin{tabular}{|c|c|c|c|c|}
        Verb & $N$ for drift & $N$ for selection & $s$ & $p$-value \\
        awake & 1150 & 3460 & 0.023 & 0.006\\ 
        blow & 26200 & 26700 & 0.00064 & 0.87\\ 
        burn & 1040 & 3790 & -0.013 & 0.004\\ 
        catch & 12500 & 47200 & 0.03 & 0\\ 
        cleave & 374 & 399 & -0.0074 & 0.41\\ 
        creep & 11200 & 13400 & 0.016 & 0.14\\ 
        dive & 1120 & 1550 & 0.012 & 0.046\\ 
        dream & 3030 & 5060 & -0.0061 & 0.074\\ 
        dwell & 2570 & 4140 & -0.02 & 0.01\\ 
        freeze & 14800 & 24300 & 0.013 & 0.086\\ 
        grow & 190000 & 190000 & -0.0022 & 0.36\\ 
        hang & 3850 & 8570 & 0.0064 & 0.022\\ 
        heave & 1530 & 2210 & -0.0064 & 0.12\\ 
        hew & 9390 & 10200 & -0.0016 & 0.4\\ 
        kneel & 978 & 2630 & 0.015 & 0.02\\ 
        lean & 1840 & 2300 & -0.0059 & 0.28\\ 
        leap & 3010 & 5730 & 0.007 & 0.02\\ 
        learn & 5100 & 8270 & -0.0052 & 0.072\\ 
        shake & 22800 & 32600 & 0.021 & 0.026\\ 
        shear & 1550 & 2980 & -0.011 & 0.008\\ 
        shine & 2570 & 2540 & -0.002 & 0.78\\ 
        slay & 12100 & 162000 & -0.026 & 0.01\\ 
        slink & 3280 & 5200 & -0.027 & 0.004\\ 
        smell & 487 & 863 & -0.015 & 0.018\\ 
        sneak & 2750 & 4340 & 0.056 & 0\\ 
        spell & 890 & 3350 & -0.014 & 0.004\\ 
        spill & 624 & 7300 & -0.02 & 0\\ 
        spoil & 1860 & 1910 & 0.0035 & 0.43\\ 
        strew & 295 & 1110 & 0.031 & 0.006\\ 
        string & 12100 & 20400 & 0.0086 & 0.1\\ 
        strive & 3730 & 4800 & -0.021 & 0.008\\ 
        swell & 618 & 828 & 0.01 & 0.1\\ 
        wake & 476 & 1350 & 0.025 & 0\\ 
        weave & 2450 & 2850 & -0.0081 & 0.12\\
    \end{tabular}
\end{center}

\subsection*{40-year bins}

\begin{center}
    \begin{tabular}{|c|c|c|c|c|}
        Verb & $N$ for drift & $N$ for selection & $s$ & $p$-value \\
        bend & 12900 & 196000 & 0.014 & 0.008\\
        bet & 748 & 3900 & 0.025 & 0.008\\
        bite & 7370 & 38000 & 0.034 & 0\\ 
        blend & 5020 & 5030 & -0.00054 & 0.98\\ 
        build & 18600 & 26800 & 0.0082 & 0.27\\
        fit & 1880 & 2450 & 0.027 & 0\\ 
        glide & 70600 & 77800 & -0.0024 & 0.74\\ 
        knit & 2120 & 10400 & -0.017 & 0.13\\ 
        light & 366 & 2180 & 0.016 & 0.014\\ 
        pat & 3470 & 67700 & -0.03 & 0\\ 
        plead & 6110 & 16700 & 0.0099 & 0.054\\
        quit & 966 & 1460 & 0.049 & 0\\ 
        sit & 200 & 200 & 0.17 & 0.99\\ 
        slide & 5730 & 36500 & 0.028 & 0\\ 
        speed & 487 & 443 & -0.014 & 0.21\\ 
        spit & 1580 & 27000 & 0.017 & 0.004\\ 
        thrust & 17100 & 26400 & 0.054 & 0.008\\ 
        tread & 9620 & 26800 & -0.033 & 0\\ 
        wed & 513 & 702 & 0.035 & 0.002\\ 
        wet & 1340 & 1620 & 0.0039 & 0.16  \\
    \end{tabular}
\end{center}

\begin{center}
    \begin{tabular}{|c|c|c|c|c|}
        Verb & $N$ for drift & $N$ for selection & $s$ & $p$-value \\
        awake & 1810 & 13500 & 0.017 & 0.022\\ 
        blow & 14500 & 15200 & 0.00087 & 0.9\\ 
        burn & 758 & 5850 & -0.012 & 0.024\\ 
        catch & 18500 & 101000 & 0.023 & 0.004\\ 
        cleave & 353 & 408 & -0.0077 & 0.52\\ 
        creep & 16100 & 16300 & 0.00068 & 0.98\\ 
        dive & 700 & 1130 & 0.012 & 0.12\\ 
        dream & 2780 & 5950 & -0.0052 & 0.17\\ 
        dwell & 2670 & 3730 & -0.022 & 0.018\\ 
        freeze & 20200 & 86700 & 0.014 & 0.058\\ 
        grow & 112000 & 110000 & -0.0028 & 0.5\\ 
        hang & 3720 & 28400 & 0.0059 & 0.028\\ 
        heave & 1210 & 4750 & -0.0083 & 0.076\\ 
        hew & 8230 & 11300 & -0.0021 & 0.31\\ 
        kneel & 965 & 3050 & 0.012 & 0.16\\ 
        lean & 1640 & 3080 & -0.0082 & 0.21\\ 
        leap & 2410 & 20700 & 0.0074 & 0.012\\ 
        learn & 4610 & 10100 & -0.0046 & 0.16\\ 
        shake & 24700 & 69000 & 0.025 & 0.006\\ 
        shear & 727 & 8120 & -0.013 & 0.01\\ 
        shine & 1920 & 1920 & -0.00042 & 0.98\\ 
        slay & 16000 & 461000 & -0.028 & 0.014\\ 
        slink & 3860 & 4260 & -0.026 & 0.004\\ 
        smell & 290 & 599 & -0.016 & 0.1\\ 
        sneak & 3790 & 4200 & 0.059 & 0.004\\ 
        spell & 497 & 2340 & -0.014 & 0.044\\ 
        spill & 385 & 10600 & -0.019 & 0\\ 
        spoil & 1260 & 1230 & 0.0034 & 0.58\\ 
        strew & 300 & 656 & 0.03 & 0.016\\ 
        string & 25700 & 59300 & 0.0053 & 0.22\\ 
        strive & 3990 & 4020 & -0.022 & 0.022\\ 
        swell & 565 & 804 & 0.0081 & 0.36\\ 
        wake & 506 & 2030 & 0.022 & 0.008\\ 
        weave & 1350 & 1830 & -0.0087 & 0.27\\
    \end{tabular}
\end{center}

\newpage

\section{Sets of words used in RAE reform detections}
\label{app:spanish}

\subsubsection*{Old spellings of words in set (A): \textless ss\textgreater to \textless s\textgreater}
\textit{assegurar}, \textit{assentar}, \textit{assentir}, \textit{assunto}, \textit{confessar}, \textit{diesse}, \textit{essa}, \textit{esse}, \textit{essencia}, \textit{esso}, \textit{estuviesse}, \textit{fuesse}, \textit{gustasse}, \textit{hiciesse}, \textit{passar}, \textit{pudiesse}, \textit{quisiesse}, \textit{tassar}, \textit{tuviesse}, and \textit{usasse}.

\subsubsection*{Old spellings of words in set (B): \textless x\textgreater to \textless j\textgreater}
\textit{abaxo}, \textit{baxar}, \textit{baxo}, \textit{bruxa}, \textit{bruxería}, \textit{caxa}, \textit{conduxo}, \textit{debaxo}, \textit{dexar}, \textit{dibuxar}, \textit{dibuxo}, \textit{dixo}, \textit{enxuto}, \textit{exe}, \textit{exemplo}, \textit{exercer}, \textit{exercicio}, \textit{exército}, \textit{floxo}, \textit{fluxo}, \textit{fixar}, \textit{fixo}, \textit{quexa}, \textit{roxo}, \textit{texa}, and \textit{traxo}.

\subsubsection*{Old spellings of words in set (C): \textless y\textgreater to \textless i\textgreater}
\textit{aceyte}, \textit{aceytuna}, \textit{afeyte}, \textit{amaynar}, \textit{ayre}, \textit{bayle}, \textit{deleyte}, \textit{deydad}, \textit{estoyco}, \textit{frayle}, \textit{gayta}, \textit{heroyco}, \textit{layco}, \textit{oyga}, \textit{peyne}, and \textit{reyna}.

\subsubsection*{Old spellings of words in set (D.1): Word-final tonic syllable becoming accentuated}
\textit{accion}, \textit{alacran}, \textit{algun}, \textit{almacen}, \textit{atencion}, \textit{bailarin}, \textit{cancion}, \textit{capitan}, \textit{comun}, \textit{corazon}, \textit{estacion}, \textit{jardin}, \textit{latin}, \textit{nacion}, \textit{ningun}, \textit{opcion} \textit{razon}, \textit{recien}, \textit{region}, \textit{relacion}, \textit{segun}, \textit{Serafin}, \textit{situacion}, \textit{tambien}, and \textit{union}.

\subsubsection*{Old spellings of words in set (D.2): Non word-final tonic syllable losing its accent}
\textit{abdómen}, \textit{álguien}, \textit{Cármen}, \textit{certámen}, \textit{cólon}, \textit{crímen}, \textit{desórden}, \textit{dictámen}, \textit{exámen}, \textit{gérmen}, \textit{jóven}, \textit{márgen}, \textit{órden}, \textit{orígen}, \textit{resúmen}, and \textit{volúmen}.

\section{Maximum likelihood parameters for time series in the study of Spanish spelling reforms}

\begin{center}
    \begin{tabular}{|c|c|c|c|c|c|c|}
    Time series & T & $N$ before & $s$ before & $N$ after & $s$ after & $p$-value \\
    (A) & 1775 & 5.0 &  -0.008 & 224 & -0.37 & <0.002 \\
     & 1805 & 60 &  -0.59 & 292 & -0.10 & 0.024 \\
    (B) & 1825 & 14 &  -0.29 & 169 & -0.072 & <0.002\\
    & 1810 & 17 & 0.003 & 33 & -2.2 & 0.036 \\
    & 1840 & 2530 & -1.0 & 154 & -0.02 & 0.026 \\
    (C) & 1825 & 9.3 & -0.30 &  121 & -0.061 & <0.002\\
    (D1) & 1890 & 16 & -0.050 & 427 & -0.21 & <0.002\\
    (D2) & 1890 & 85 & 0.13 & 249 & -0.53 & <0.002\\
    \end{tabular}
\end{center}


\end{document}